\documentclass[journal]{IEEEtran}
\usepackage{pdfpages}
\usepackage{float}

\usepackage{cite}

\ifCLASSINFOpdf
\else
\fi
\usepackage{amsmath}
\usepackage{algorithmicx,algorithm}

\usepackage{booktabs}
\usepackage{multirow}
\usepackage{array}
\begin{document}
%
\title{Polarimetric SAR Image {Semantic} Segmentation with 3D Discrete Wavelet Transform and Markov Random Field}
%
%
%
%

\author{
Haixia~Bi, Lin~Xu,~Xiangyong~Cao, Yong Xue, and Zongben Xu
\thanks{This work was supported by the China Postdoctoral Science Foundation funded project (2018M643655), the Fundamental Research Funds for the Central Universities, and the China NSFC project under contract 61806162 and 61906151. \emph{(Corresponding author: Lin Xu.)}

H.~Bi is with Faculty of Engineering, University of Bristol, Bristol
BS8 1UB, United Kingdom (e-mail: bhxwzq@163.com.)

L.~Xu is with the Institute of Artificial Intelligence, Shanghai Em-Data Technology Co., Ltd., Shanghai 200000, China  (lin.xu5470@gmail.com).

Y.~Xue is with the School of Environment Science and Spatial Informatics,
China University of Mining and Technology, Xuzhou, Jiangsu 221116, China,
and also with the School of Electronics,
Computing and Mathematics, University of Derby,
Derby DE22 1GB, United Kingdom (e-mail: y.xue@derby.ac.uk)

X.~Cao, and Z.~Xu are with Institute for Information and System Sciences, Xi'an Jiaotong University,
Xi'an, China (caoxiangyong45@gmail.com, and zbxu@mail.xjtu.edu.cn).
}}

%
%

\markboth{FINAL SUBMISSION TO IEEE TRANSACTIONS ON IMAGE PROCESSING}%
{Shell \MakeLowercase{\textit{et al.}}: Bare Demo of IEEEtran.cls for IEEE Journals}
%



\maketitle
\begin{abstract}
Polarimetric synthetic aperture radar (PolSAR) image segmentation
is currently of great importance in
image processing for remote sensing applications.
However, it is a challenging task due to two main reasons.
Firstly, the label information is difficult to acquire
due to high annotation costs.
Secondly, the speckle effect embedded in the PolSAR imaging process
remarkably degrades the segmentation performance.
To address these two issues, we present a contextual PolSAR image
semantic segmentation method in this paper.
With a newly defined channel-wise consistent feature set as input,
the three-dimensional discrete wavelet transform (3D-DWT) technique
is employed to extract discriminative multi-scale features
that are robust to speckle noise.
Then Markov random field (MRF) is further applied to enforce label smoothness
spatially during segmentation.
By simultaneously utilizing 3D-DWT features
and MRF priors for the first time,
contextual information is fully integrated during the segmentation
to ensure accurate and smooth segmentation.
To demonstrate the effectiveness of the proposed method,
we conduct extensive experiments on {three} real benchmark PolSAR image data sets.
Experimental results indicate that the proposed method
achieves promising segmentation accuracy
and preferable spatial consistency using
a minimal number of labeled pixels.
\end{abstract}


%
\IEEEpeerreviewmaketitle

\section{Introduction}
%
%
%
%

\IEEEPARstart{T}{he} past few decades witnessed significant progress in polarimetric synthetic aperture radar (PolSAR) theories and applications.
With the flourish of airborne and spaceborne PolSAR systems,
massive high-resolution PolSAR images are collected nowadays. PolSAR image {semantic} segmentation, the process of dividing
PolSAR images into different terrain categories,
has received increasing attentions
in modern civil and military applications. A great variety of state-of-the-art PolSAR image segmentation approaches have been proposed in recent years \cite{Lee1994,Cloude1997,Lee1999,Freeman1998,HeWL2013,DeepLearning,TexfeaMRF2016,
TGRS_Bi,TGRS_Bi2,TGRS_Bi3,hong2019,yao2019,Doulgeris2015,IGARSS2019,
Kong1988,Pottier1991,Antropov2014,Fukuda2001,WishartDBN,WishartDSN,
Fourcom2005,Lee_Kdistri1994,Doulgeris2011,HouBin2016,WuMRF2008,
hong2020,TMF2014,Wishart1994,SAESDPL,XuFact,XuAnis,rasti2020,cospace2019,SSA2016,XuDeori2005,Cao2020}.

However, despite the rapid progress,  there are still some challenges
in PolSAR image {semantic} segmentation. Firstly,
it is well acknowledged that high-quality annotated data is difficult to acquire, which is extremely severe in PolSAR data processing task. Annotation of the PolSAR image is not only labor-intensive and time-consuming, but also demanding specialty-oriented knowledge and skills, which leads to the scarcity of ground truth {semantic} information.
Secondly, PolSAR images are susceptible to speckle noise caused by
the coherent imaging mechanism of PolSAR systems, which dramatically degrades the quality of PolSAR images and the following segmentation performance.

A great number of methods have been proposed to address the above two issues \cite{HeWL2013,TexfeaMRF2016,DeepLearning,TGRS_Bi,TGRS_Bi2,TGRS_Bi3,
	WuMRF2008, HouBin2016,Doulgeris2015,rasti2020}.
One primary strategy is using more representative features,
such as wavelet analysis \cite{HeWL2013},
Gabor filtering\cite{TexfeaMRF2016} and
convolutional neural network (CNN) \cite{DeepLearning,TGRS_Bi2,TGRS_Bi3,rasti2020} etc.
Compared with the original polarimetric indicators,
these extracted features incorporate local semantic information
by convolutional operations over a specified region,
yielding higher spatial consistency,
which can effectively depress speckle noise.
In addition, a group of discriminative and representative features
are beneficial for the classifier to approach the decision boundary,
and therefore relieves the reliance of classifiers on annotations.
Another strategy dedicates to enforcing
the local consistency on pixel labels.
Typical approaches include the over-segmentation technique
utilized in preprocessing step
and graph-based optimization
executed as a post-processing process.
Over-segmentation technique divides the whole PolSAR image
into small homogeneous patches or superpixels which are considered as an integral part in the following classifier learning
and labeling process \cite{WuMRF2008,HouBin2016}.
Pixels in each patch share the same label during the segmentation.
Graph-based optimization incorporates label smoothness priors
into the segmentation task
by solving a maximum a posterior (MAP) problem
\cite{TexfeaMRF2016,Doulgeris2015,TGRS_Bi,TGRS_Bi2}.
With the integrated contextual consistency priors,
the speckle noise in PolSAR images can be effectively depressed,
producing segmentation results with higher classification accuracy
and better spatial connectivity.

Both the above two strategies can effectively
promote the performance of PolSAR image {semantic} segmentation.
However, most of the existing PolSAR image segmentation methods
usually consider only one of them.
It is worth noting that
although the prevailing deep neural networks
are capable of learning discriminative and data-driven features,
their performance is subject to
the availability of large amounts of annotated data which
require great efforts of experienced human annotators.
In addition, the time consumption of deep neural networks
is usually high.
Therefore, to make the most advantage of contextual {semantic} information,
{and reduce the time consumption and the reliability on annotations,}
we propose a new PolSAR image {semantic} segmentation pipeline in this work.  The main inspiration is to integrate multi-scale texture features and {semantic} smoothness priors into a principled framework.
Specifically, we use the three-dimensional discrete wavelet transform (3D-DWT) to extract multi-scale PolSAR features.
By incorporating the connections between features
in the third dimension,
the 3D-DWT features are more representative
than traditional 2-dimensional (2D) texture features, yet
could be extracted without complex learning process.
Then,  we utilize Markov Random Field (MRF) to refine the segmentation results by enforcing label smoothness and the alignment of label boundaries with image edges. It can effectively counteract speckle noise
while achieving better spatial connectivity and classification accuracy.
The main contributions of this work are summarized as follows:

(1)  We formulate the spatial {semantic} features using 3D-DWT
and label smoothness priors using MRF into a principled framework.
To the best of our knowledge, this is the first work
that simultaneously incorporates 3D-DWT and MRF
into PolSAR image {semantic} segmentation.

(2) Different from traditional MRF models
which only incorporate label smoothness priors,
our defined MRF not only encourages the spatial consistency
but also enforces the alignment of label boundaries with image edges.
{{Belief propagation (BP) algorithm
is employed to optimize the MRF model due to its fast convergence.}}

(3)  To evaluate the performance of the proposed
method, we conduct extensive experiments on {three} real benchmark
PolSAR images. Experimental results show the
advantages of our proposed method compared with the other eight competing PolSAR image {semantic} segmentation methods.

The remainder of this paper is organized as follows.
We introduce the preliminaries, including the related works of PolSAR
image {semantic} segmentation,
3D-DWT and graph model in Section II.
Then, we describe the proposed method,
including the raw polarimetric indicators,
3D-DWT feature extraction,
SVM-MRF based supervised classification model
and the model optimization in Section III.
Section IV reports experimental results
on {three} real benchmark PolSAR images.
Conclusions and future works are discussed in Section V.

\section{Preliminaries}
\subsection {Related works}
Considering the foundations of current PolSAR image {semantic} segmentation methods, we broadly divide them into three categories: scattering mechanism-based methods,
statistics-based methods and machine learning-based methods,
which will be detailed below.

\subsubsection {Scattering mechanism-based methods}
This category of methods conduct segmentation based on diverse
polarimetric target decomposition methods, from which
various terrain classes can be derived with explicit physical meanings.
The most classical scattering mechanism-based method is
the $H/\bar{\alpha}$ segmentation approach proposed by
Cloude and Pottier \cite{Cloude1997}.
In this method, eigenanalysis is firstly performed on
polarimetric coherency matrix, constructing a 2-D feature plane
using the extracted scattering entropy \emph{H}
and average scattering angle $\bar{\alpha}$.
The $H/\bar{\alpha}$ plane is then divided into eight subspaces
which represent diverse scattering mechanisms.
Based on this subspace division,
each pixel can be projected to one of the eight primary zones,
determining its terrain category finally.
Other commonly employed target decomposition methods include
Freeman decomposition\cite{Freeman1998} and
four-component decomposition\cite{Fourcom2005} {etc}.

\subsubsection {Statistics-based methods}
With the development of polarimetric theories,
researchers discovered that polarimetric data comply with
certain statistical laws.
Lee \emph{et al.} proposed the complex Wishart distribution
for both coherency matrix and covariance matrix.
Based on this hypothesis, a Wishart distance was defined
to reveal the similarity of a pixel
to a certain terrain clustering center\cite{Wishart1994,Lee1994}.
Lee \emph{et al.} innovatively combined polarimetric distribution
with target decomposition theories\cite{Lee1999}.
Specifically, an initial segmentation
is firstly conducted using $H/\bar{\alpha}$ decomposition,
and then the segmentation map is iteratively updated
based on Wishart distance.
In addition to the complex Wishart distribution,
other polarimetric statistical hypotheses as
$K$-distribution\cite{Lee_Kdistri1994}
and $\mathcal{U}$-distribution \cite{Doulgeris2015}
were also brought forward and employed in PolSAR image {semantic}
segmentation task \cite{Doulgeris2011, TGRS_Bi, TGRS_Bi2}.

\begin{figure*}[htb]
			\vspace{-.2cm}
	\begin{center}
		\includegraphics[height=6.0cm,width=13.5cm]{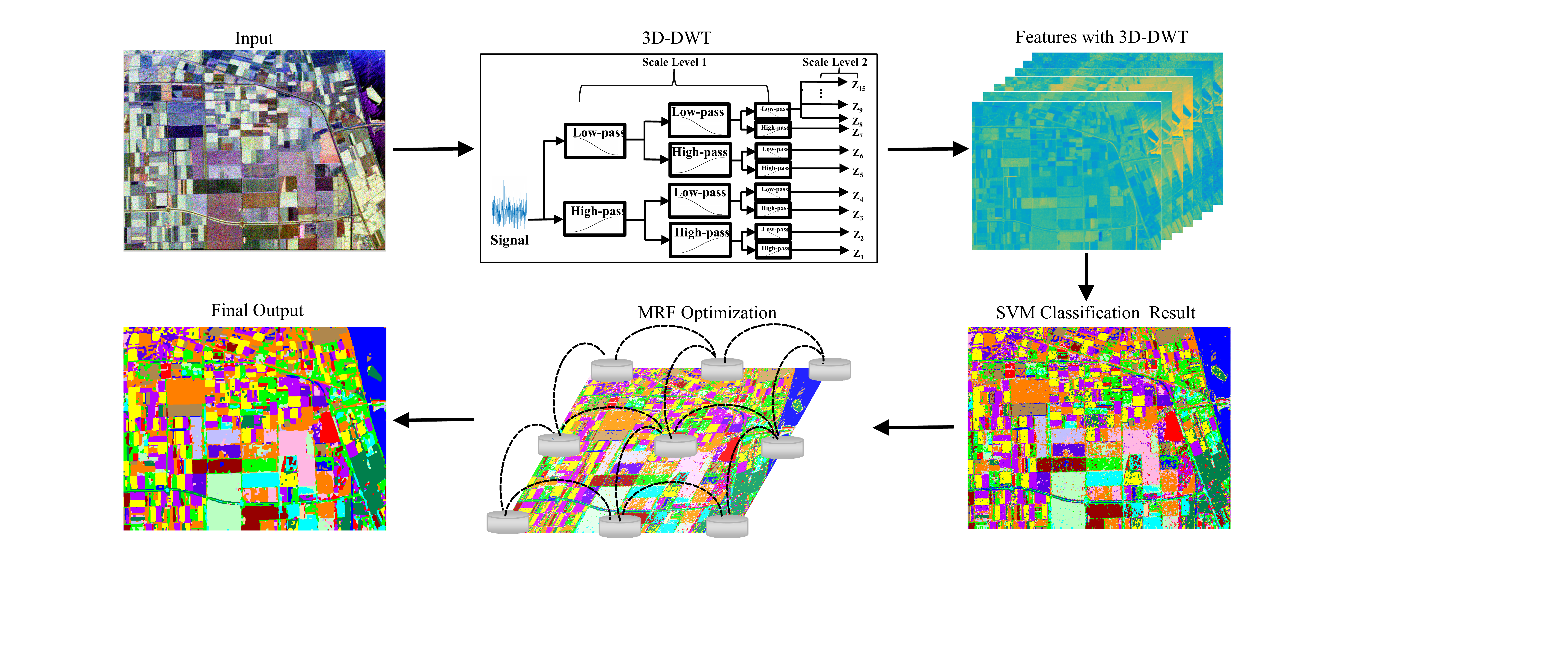}
		     		\vspace{-.3cm}
		\caption{Illustration of our proposed method. The whole pipeline includes three  steps. In step 1, based on the raw features extracted from PolSAR data,
3D-DWT is employed to generate spatial texture features.
In step 2, SVM is used to generate class probabilities and a preliminary classification map.
In step 3,  MRF  optimization is conducted to refine the classification map.}
	\end{center}
		\vspace{-.6cm}
\end{figure*}

\subsubsection {Machine learning-based methods}
Machine learning approach has dominated
the PolSAR image  {semantic} segmentation task in recent years.
Bayesian classification method was firstly introduced to
PolSAR image segmentation by Kong \emph{et al}\cite{Kong1988}.
Pottier \emph{et al.}\cite{Pottier1991}
and Antropov \emph{et al.}\cite{Antropov2014} applied neural network in
PolSAR image {semantic} segmentation.
Support vector machine (SVM) was extensively employed
in this task \cite{Fukuda2001,cospace2019,HeWL2013,TexfeaMRF2016},
producing desirable segmentation results
due to its elaborate optimization architecture and good generalization ability.
Most recently, the advent of deep learning techniques
provided a new way for PolSAR image segmentation.
Zhou \emph{et al.}\cite{DeepLearning} and Bi \emph{et al.} \cite{TGRS_Bi2,TGRS_Bi3} exploited  convolutional neural network (CNN)
to extract hierarchical polarimetric features.
Deep belief network was applied to PolSAR image segmentation
by Liu \emph{et al} \cite{WishartDBN}.
Zhang \emph{et al.} \cite{SSA2016} utilized stacked sparse autoencoder
to learn deep spatial sparse features of PolSAR.
Jiao \emph{et al.} \cite{WishartDSN} designed a
PolSAR image segmentation method based on deep stacking network.
Chen \emph{et al.} \cite{SAESDPL} proposed a
novel semicoupled projective dictionary pair learning method (DPL)
with stacked auto-encoder (SAE) for PolSAR image classification.
The SAE is utilized to obtain hierarchical features,
while SDPL is employed to reveal
the intrinsic relationship between different features
and perform classification.
MRF technique has also been employed
in PolSAR image segmentation task.
Wu \emph{et al.} \cite{WuMRF2008} proposed a region-based
PolSAR image segmentation method which combines MRF
and Wishart distribution. Liu \emph{et al.} \cite{TMF2014}
designed a PolSAR image segmentation method
based on Wishart triplet Markov field (TMF).
Masjedi \emph{et al.}\cite{TexfeaMRF2016} incorporated
texture features and contextual information realized
by MRF technique in PolSAR image segmentation.
Doulgeris \emph{et al.}\cite{Doulgeris2015} put forward an automatic
MRF segmentation algorithm based on $\mathcal{U}$-distribution for PolSAR images.
In addition, these are new trends of multi-dimentional PolSAR image analyses.
Xu et al. \cite{XuFact,XuAnis} recently proposed novel POLSAR image analysis methods which exploit spatial and aspect dimensions jointly with polarimetric features.

Investigating the theories and performances of
these PolSAR image {semantic} segmentation methods,
we can discover that: 1) Methods falling into the first two categories
are simple, fast, and physically interpretable,
however, the segmentation results are coarse and imprecise,
making them only suitable for preliminary analysis of PolSAR data.
2) Machine learning-based methods, especially the supervised methods,
achieve better segmentation performance than the former two
categories of methods. However, they rely heavily on large amounts
of annotated data, even for the most popular deep learning-based methods.
In addition, the deep learning-based methods usually consume a
long running time.
3) By incorporating contextual information,
MRF effectively improves segmentation performance
and enforces label smoothness.

Inspired by the above analysis, we aim to explore
more discriminative polarimetric features and fully utilize
the contextual information of PolSAR data in this work.
Specifically, 3D-DWT is firstly employed
to extract multi-scale contextual polarimetric features.
Moreover, a newly defined MRF model is further applied to refine the
segmentation map, which not only enforces label smoothness,
but also encourages the alignment of label boundaries with image edges.

\subsection {Wavelet analysis}
The wavelet transform (WT) \cite{DWT3D} is a popular mathematical tool
for time-frequency analysis.
Through dilating and shifting operation,
sets of wavelets are generated from the original mother wavelet,
which can be used to analyze different proportions of signals.
Specific for image processing, the high-frequency proportion
represents small-scale details, such as the edges of images,
while the low-frequency proportion corresponds to the
smoothing part of images.

The advent of multi-resolution analysis (MRA) \cite{Mallat}
dramatically promoted the practical application of wavelet analysis.
MRA aims at constructing a set of orthogonal wavelet bases
which can infinitely approximate $L^2(R)$ space in the frequency domain.
This provides an efficient way to analyze various
proportions of signals.
Owing to its outstanding localization characteristics
and multi-resolution analysis ability,
wavelet analysis has been extensively applied in multiple fields, including
image processing, audio analysis, and theoretical physics {etc}\cite{HeWL2013,TexfeaMRF2016}.
The commonly utilized wavelets include Haar wavelet, Daubechies wavelet,
and Morlet wavelet {etc}.

\subsection {Graph Model in Image Segmentation}
Graph model is one of the most important models in image processing
owing to its solid mathematical foundation \cite{GraphBasedSemi}.
It provides a convenient approach to depict the local consistency
between pixels in an image.
For image segmentation task,
an undirected graph $\mathcal{G} =<\mathcal{V},\mathcal{E}>$
can be established over the image,
where image node set $\mathcal{V}$
corresponds to pixels, and undirected edge set $\mathcal{E}$
represents the neighborhood relationship, i.e., similarities,
between pixels \cite{JianSun}.
Then the labels of $\mathcal{V}$ spread through
edges $\mathcal{E}$ using an optimization
function, realizing segmentation of graph $\mathcal{G}$.

If positivity and Markovianity of variables on the graph are satisfied,
the graph can be regarded as an MRF.
Image segmentation problem, with class labels as variables,
is a typical MRF solving problem.
Based on the Hammersley-Clifford theorem and Bayesian theory \cite{Gibbs},
the labeling task in MRF can be transformed
to solving a MAP problem.

\section{The proposed method}

\subsection{Overview of the Proposed Method}
In this section, we will first define some notations,
and then introduce the pipeline of the proposed method.
For a given PolSAR image, the raw PolSAR feature data
is defined as $\mathbf{X} \in {R^{H\times{W}\times{D}}}$,
where $H$ and $W$ are the height and width of the PolSAR image,
and $D$ is the dimension of the selected raw polarimetric indicator
as described in Section II.
The labeled training samples are denoted as ${(\mathbf{x}_i,\mathbf{y}_i)}_{i=1}^{n}$,
where $n \ll H\times{W}$, $\mathbf{y}_i \in \{1, \cdots K \}$.
Here, $K$ is the total number of classes.
The proposed method is designed to assign label $\mathbf{y}_i$
to each pixel $i(i \in{1,\cdots,N)}$, where $N=H\times{W}$.
We further denote $\mathbf{Y}=\{\mathbf{y}_i\}$ in the following sections.

Taking the Flevoland area data set 1 as example,
Fig.~1 illustrates the framework of the proposed method,
which formulates 3D-DWT contextual feature extraction
and SVM-MRF classification into a unified model.
Raw polarimetric indicators are firstly extracted
from the original polarimetric matrices (Section III.B).
We then implement 3D-DWT feature extraction on the raw features to
obtain multi-scale contextual features (Section III.C).
The consequently executed SVM-MRF classification method consists of
two subproblems, i.e., SVM learning-subproblem and label propagation-subproblem (Section III.D-E).
We will detail all the above components in the following paragraphs.
\vspace{-1mm}
\subsection{Polarimetric Raw Indicators}
We first define a raw polarimetric feature set as input of our proposed method.
{All the features are directly drawn from the second-order $3\times 3$
complex coherency polarimetric matrix $T$,
which are shown as:

\vspace{-2mm}
\begin{equation*}
\{\rm{SPAN}\rm,\emph{T}_{11},\emph T_{22},\emph T_{33},|\emph T_{12}|,|\emph T_{13}|,|\emph T_{23}|\}   \\ \nonumber
\end{equation*}

In the employed 7-dimensional (7D) feature set, \rm{SPAN} ($\rm{SPAN}$=$T_{11}+T_{22}+T_{33}$)
denotes the total polarimetric power.
The remaining six features indicate the intensities of
the diagonal and upper triangle elements
of complex coherency polarimetric matrix $T$.}
Since the 7D features are all transformed
from the original scattering matrix,
they represent different spectrums of polarimetric signals,
yet interrelated to each other.
Therefore, they can be considered as
diverse information channels of the terrain targets,
which makes it a natural and straightforward choice to
perform 3D-DWT on the polarimetric feature cube.

\vspace{-0.1cm}
\subsection{3D-DWT Feature Extraction}
Define $f(x)$ as a quadratic integrable function,
and $\phi(x)$ as the mother wavelet function which
satisfies the admissibility condition.
The wavelet analysis is defined as:
\begin{eqnarray} \label{CWT}
W_{\phi f}{(a,b)}=<f(x),\phi_{a,b}(x)>=\int
f(x)\frac{1}{\sqrt{a}}\phi{(\frac{x-b}{a})}dx ,
\end{eqnarray}
where $\phi_{a,b}(x)$ is wavelet basis function
which is obtained from mother wavelet $\phi(x)$
through dilating and shifting operation.
$a$ is called scaling parameter and $b$ is called shifting parameter.

Continuous wavelet transform (CWT) is greatly suitable
for data analysis due to the detailed description of signals.
However, CWT is computation resource costly and
information redundant due to the similarity
between wavelet components, which makes it inapplicable
in practical applications. To address this problem,
CWT is transformed to discrete wavelet transform (DWT)
through discretization on both time domain and
frequency domain.

Generally, dyadic discretization is carried out on
scaling parameter $a$ as 1, $a_0$, $a_0^2$, $a_0^3$, ...,
$a_0^j$. Under scale level $a_0^j$, taking $b_0$
as the shifting step, $b$ is discretized as 0,
$a_0^jb_0$, $2a_0^jb_0$, ..., $ka_0^jb_0$.
Then the DWT is written as:
\begin{eqnarray} \label{DWT}
W_{m,n}^{\phi}(f)=<f(x),\phi_{m,n}(x)>=\int f(x)\phi_{m,n}(x)dx ,
\end{eqnarray}
where $\phi_{m,n}(x)=a_0^{-m/2}\phi((x-n{b_0}{a_0^m})/{a_0^m})$,
$a_0$ and $b_0$ are dyadic scale parameter and shifting parameter, respectively.

According to multi-resolution analysis (MRA), function $f(x)$
can be approximated using a linear combination of
scaling function $\psi(x)$ and wavelet function $\phi(x)$
which represent low-frequency approximation and
high-frequency detail respectively.
A discrete signal $f[n]$ can be approximated by
\begin{eqnarray} \label{DisSig}
f(n)&=&\frac{1}{\sqrt M}\sum_{k}C_{\psi}[j_0,k]\psi_{j_0,k}[n] \nonumber \\
&+& \frac{1}{\sqrt M}\sum_{j=j_0}^{\infty}\sum_k D_{\phi}[j,k]\phi_{j,k}[n],
\end{eqnarray}
where $j_0$ is any starting scale, $C_{\psi}[j_0,k]$ is
called scaling coefficient, and $D_{\phi}[j,k]$ is
called wavelet coefficient. In DWT, their values are:
\begin{eqnarray} \label{ScalWavCoef}
C_{\psi}[j_0,k]=\frac{1}{\sqrt M}\sum_{n}f(n)\psi_{j_0,k}(n), \nonumber \\
D_{\phi}[j,k]=\frac{1}{\sqrt M}\sum_{n}f(n)\phi_{j,k}(n),
\end{eqnarray}
where $f[n]$, $\psi_{j_0,k}[n]$ and $\phi_{j,k}[n]$ are discrete functions defined on $[0,M-1]$, containing totally $M$ points.
	\vspace{-.2cm}
\begin{figure}[htb]
	\begin{center}
		\includegraphics[height=4.0cm,width=7.2cm]{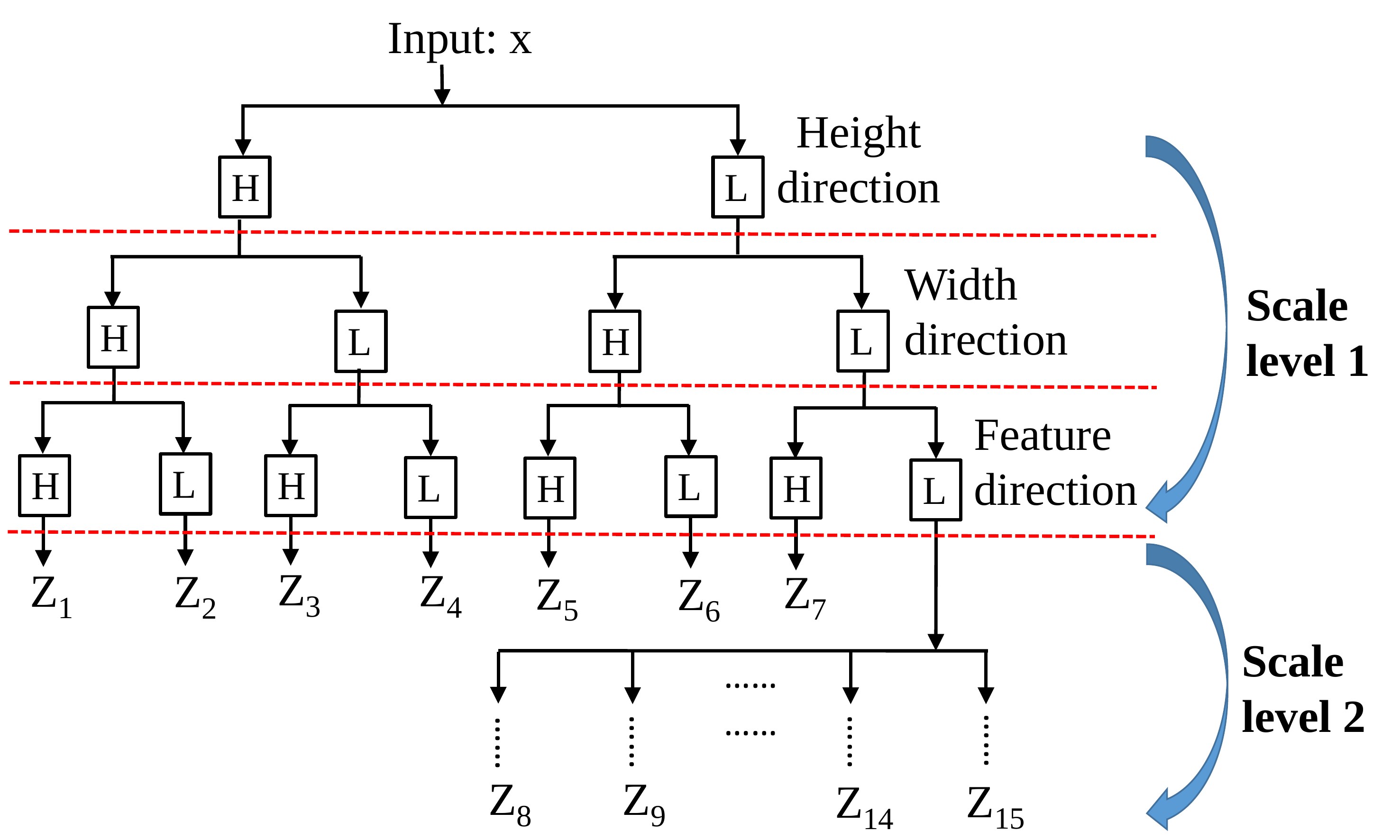}
		\caption{3D-DWT feature extraction.
The PolSAR data cube is decomposed in two scale levels.
At each scale level, convolutions with all combinations of
high-pass and low-pass filters in three dimensions
yield eight different PolSAR data cubes.
Then, the PolSAR data cube filtered by the low-pass filters
in three dimensions is further decomposed in the next scale level.}
	\end{center}
	\vspace{-.3cm}
\end{figure}

In this work, we apply the 3D-DWT technique to polarimetric data cube,
which encodes the contextual information to different scales.
It is noteworthy that 3D-DWT can be achieved
by applying 1D-DWTs to each of the three dimension.
For this application, MRA decomposes the original signal
into low-frequency part and high-frequency part,
and then continually decomposes
low-frequency part while keeping the high-frequency part unchangeable.
In practice, scaling and wavelet functions are achieved
using a filter bank $[L,H]$, where $L$ are low-pass filters
with coefficients $C_{\psi}[j,k]$
and $H$ are high-pass filters with coefficients $D_{\phi}[j,k]$.
The Haar wavelet is employed in this work, where $C_{\psi}[j,k]=(1/\sqrt{2},1/\sqrt{2})$,
and $D_{\phi}[j,k]=(-1/\sqrt{2},1/\sqrt{2})$.
The decomposition structure as shown in Fig.2 is
employed in this work. After the 3D-DWT decomposition,
15 sub-cubes ${Z}_1, {Z}_2,...,{Z}_{15}$ are generated.
It should be noted that the down-sampling step is left out
in the process, and thus each sub-cube has the same size
with the original polarimetric data cube.
For pixel $(i,j)$, the 3D-DWT wavelet coefficients
can be concatenated to below feature vector form.
\begin{eqnarray} \label{FV}
{z}(i,j) = ({Z}_1(i,j,\cdot), {Z}_2(i,j,\cdot),
\cdots, {Z}_{15}(i,j,\cdot)),
\end{eqnarray}

To further enforce spatial consistency,
a $3\times 3$ square mean filter is applied to
the absolute values of 3D-DWT wavelet coefficients following:

\begin{eqnarray} \label{ScalWavCoef}
\bar{Z}_k(i,j) = \frac{1}{9} \sum_{m=i-1}^{i+1} \sum_{n=j-1}^{j+1}
|{Z}_k(m,n,\cdot)|, k=1,2,\cdots,15,
\end{eqnarray}

We denote ${\mathbf{Z}} \in {R^{H\times{W}\times{15D}}}$
as the final concatenated feature vector in the following sections.
{With the 7D raw feature set (described in Section III.\emph{B})
as input, the 3D-DWT feature extractor outputs 105-dimensional
texture features, acting as input for the SVM classifier.}

\subsection{SVM-MRF Classification Model}

Given the extracted 3D-DWT features as input, we next introduce
our proposed SVM-MRF PolSAR image {semantic} segmentation model.
It aims to estimate class label $\mathbf{Y}$
with observation $\mathbf{Z}$ (we use the 3D-DWT features $\mathbf{Z}$ instead of the original raw features $\mathbf{X}$).
To serve this propose, we define an energy function given by:
\begin{eqnarray} \label{TotalLoss}
E=E_d+E_s,
\end{eqnarray}
where $E_d$ indicates the SVM data loss term,
and $E_s$ denotes the label smoothness term.
The dense class labels $\mathbf{Y}$ can be predicted
by minimizing this energy function,

\subsubsection{SVM data loss term}
This term is designed to predict class labels based
on the contextual 3D-DWT features,
which follows the multi-class probabilistic SVM \cite{LIBLINEAR}
as
\begin{eqnarray} \label{DataLoss}
E_d=-\sum_{i=1}^{N}log{P(\mathbf{y}_i|\mathbf{z}_i)},
\end{eqnarray}
where $P(\mathbf{y}_i|\mathbf{z}_i)$ denotes the probabilities
of pixel $i$ belonging to class $\mathbf{y}_i$ with feature $\mathbf{z}_i$.
Minimizing this term encourages that the class label
with higher output probability will be preferred.
The larger the probability of a pixel belonging to a certain class,
the more probable that the pixel is assigned with
the corresponding label.

\subsubsection{Label smoothness term}
The label smoothness term is defined
to enforce the smoothness of estimated class label map
and alignment of class label boundaries with image edges,
which is defined as
\begin{eqnarray} \label{SmoothnessCost}
E_s=\alpha_s\sum_{i=1}^N\sum_{j\in \mathcal{N}(i)}\mathcal{S}_{ij},
\end{eqnarray}
where $\alpha_s$ is the label smoothness factor.
$\mathcal{N}(i)$ is the neighboring pixel set of pixel $i$.
$\mathcal{S}_{ij}$ is defined as
\begin{eqnarray} \label{SmoothnessFunc}
\mathcal{S}_{ij}=|\mathbf{y}_i-\mathbf{y}_j|\exp(-\frac{\|\mathbf{v}_{i}-\mathbf{v}_{j}\|_2^2}{2\sigma}),
\end{eqnarray}
where $\mathbf{v}_{i}$ is a feature vector located at pixel $i$,
which should be chosen as the features whose values
significantly change across edges in image.
We take $\mathbf{v}_{i}$ as the Pauli matrix components in this work.
$\sigma$ indicates the mean squared distance
between features of two adjacent pixels $i$ and $j$.
The label smoothness loss function encourages
the label boundaries to align with strong image edges.
For pixels $i$ and $j$ within flat regions, $\exp(-\frac{\|\mathbf{v}_{i}-\mathbf{v}_{j}\|_2^2}{2\sigma})$ in Eq.~(\ref{SmoothnessFunc}) is large, then minimizing
$\mathcal{S}_{ij}$  will intensify the
chance that labels $\mathbf{y}_i$ and $\mathbf{y}_j$ take same class label.
However, for neighboring pixels spanning strong edges, $\exp(-\frac{\|\mathbf{v}_{i}-\mathbf{v}_{j}\|_2^2}{2\sigma})$ is smaller (or even near to zero), thus the inconsistency between the class labels of neighboring pixels $i$ and $j$ is allowable during optimization.

Summarized
from the above formulations, the final integrated energy function
can be written as
\begin{eqnarray} \label{TotalCost}
E=-\sum_{i=1}^{N}log{P(\mathbf{y}_i|\mathbf{z}_i)}+
  \alpha_s\sum_{i=1}^N\sum_{j\in \mathcal{N}(i)}\mathcal{S}_{ij},
\end{eqnarray}

\subsection{Optimization}
Label $\mathbf{Y}$ can be solved by minimizing the energy function
defined in Eq.~(\ref{TotalCost}).
We decompose this optimization problem into two subproblems,
i.e. \emph{SVM learning-subproblem} and \emph{Label propagation-subproblem}.

\subsubsection{SVM learning-subproblem}
For this subproblem, a multi-class probabilistic
SVM classifier is firstly trained using a preselected training set
based on 3D-DWT features.
Next, the learned classifier is employed to
predict pixel-wise class probabilities of the whole data set,
providing input for the label propagation-subproblem.

\subsubsection{Label propagation-subproblem}
Based on the trained classifier in SVM learning-subproblem,
this subproblem focuses on updating label $\mathbf{Y}$
while incorporating label smoothness priors.
This label assignment problem is
a combinatorial optimization problem,
which can be regarded as a standard MRF model\cite{TGRS_Bi}.
The labels $\mathbf{Y}$ over the graph constitute an MRF,
and Eq.~(\ref{TotalCost}) is the energy function on it.
The two terms of the energy function enforce
label constraints from the agreement between
SVM predictions and target class labels,
and class label smoothness, respectively.
Minimizing these two terms enforces
that the predicted class labels should be
smooth and consistent with the SVM predictions.

{Labeling problem in an MRF is a NP-hard problem.
However, optimal solution can be approximately achieved using
optimization algorithms \cite{ICM,BP2003}.
Belief propagation (BP) algorithm \cite{BP2003} iteratively
optimizes MRF using an ``up-down-left-right" message passing schedule
in a linear time complexity,
which makes the model converge very quickly.
Therefore, BP algorithm is employed
to optimize our defined MRF model in this work.

\begin{figure}[htb]
			\vspace{-.3cm}
\begin{center}
\includegraphics[height=3.8cm,width=4.2cm]{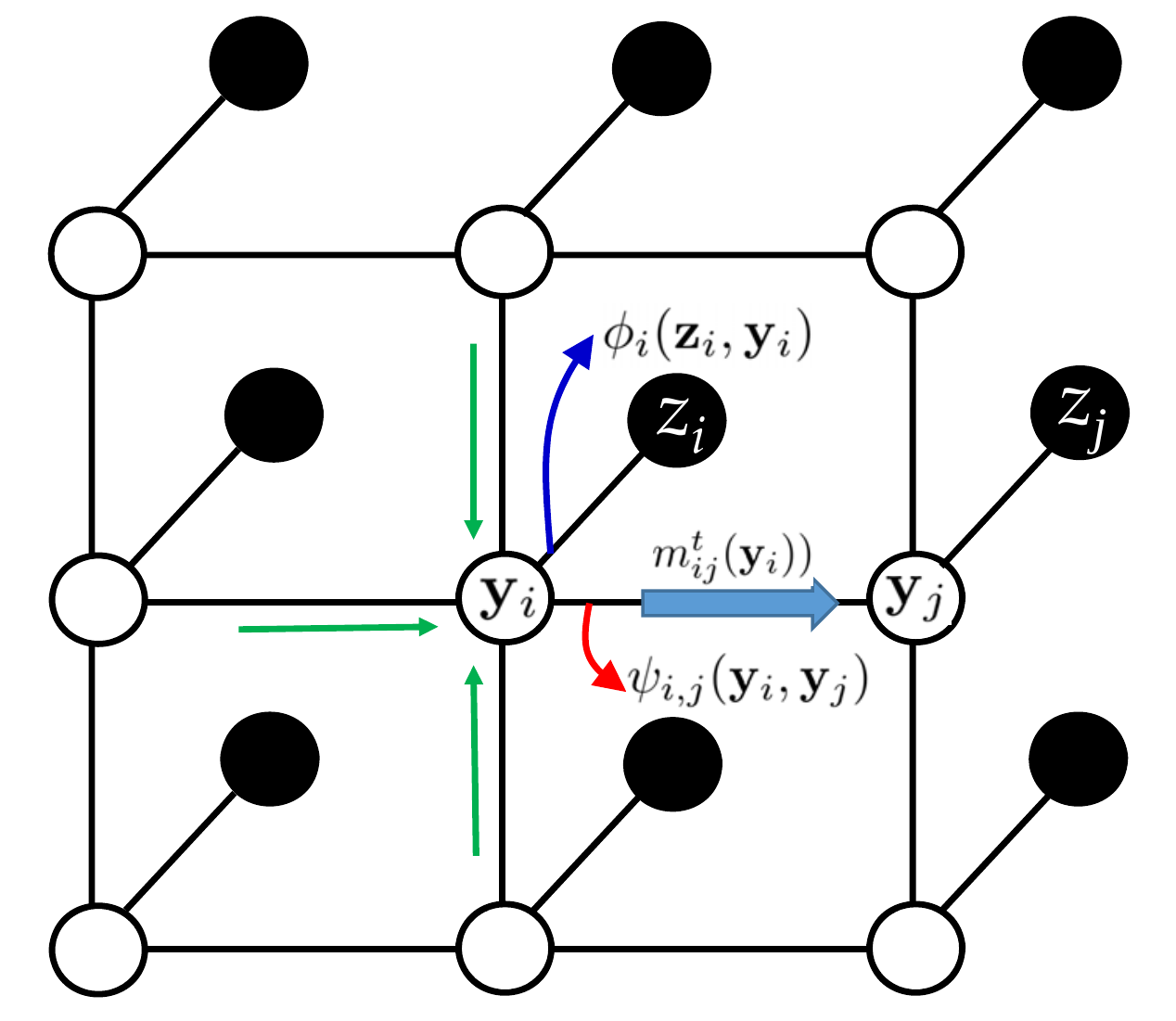}
		\vspace{-.3cm}
\caption{Message passing scheme in MRF model using BP algorithm.}
\end{center}
		\vspace{-.6cm}
\end{figure}

Figure 3 illustrates the message passing policy
of BP optimization algorithm.
Each node $i$ in the MRF model corresponds
to one pixel in the given PolSAR image,
which can be considered as a random value.
Label $\mathbf{y}_i$ is linked to its 3D-DWT contextual feature data $\mathbf{z}_i$.
Let $\phi_{i}(\mathbf{z}_i,\mathbf{y}_i)$ represent the likelihood potential of
label $\mathbf{y}_i$ given feature data $\mathbf{z}_i$,
taking the form of the multi-class probabilistic SVM loss term
$-\log P(\mathbf{y}_i=j|\mathbf{z}_i)$ [See Eq.~(8)],
$\psi_{i,j}(\mathbf{y}_i,\mathbf{y}_j)$ stand for the label smoothness potential
that encourages label contextual consistency
between neighboring pixels $i$ and $j$
with form $\alpha_s\mathcal{S}_{ij}$ [See Eq.~(9-10)].
Then at iteration $t$, the message
sent from node $i$ to its neighboring node $j\in\mathcal{N}(i)$ is given by

\vspace{-2mm}
\begin{equation} \label{MRF}
\begin{split}
\small
m_{ij}^{t}(\mathbf{y}_j) \gets
\mathop{\max}\limits_{\mathbf{y}_i}(\mathop{\sum}\limits_{k\in\mathcal{N}(i)\backslash j}m_{ki}^{t-1}(\mathbf{y}_i) -\phi_{i}(\mathbf{z}_i,\mathbf{y}_i)-\psi_{i,j}(\mathbf{y}_i,\mathbf{y}_j))
\end{split}
\end{equation}
\vspace{-4mm}

With all messages initialized as 0,
messages of each node are iteratively updated and propagated to
its neighboring nodes until convergence.
Finally, label $\mathbf{y}_i$ is obtained
by estimating the maximum belief for node $i$, which is given by

\vspace{-4mm}
\begin{equation} \label{BP}
\begin{split}
\mathbf{y}_i=\mathop{\arg\max}\limits_{\mathbf{y}_i \in \{1,...,K \}}(\mathop{\sum}\limits_{k\in\mathcal{N}(i)}m_{ki}^t(\mathbf{y}_i) -\phi_{i}(\mathbf{z}_i,\mathbf{y}_i))
\end{split}
\end{equation}
}
The open source code package MRF minimization\footnote{[Online]. Available: http://vision.middlebury.edu/MRF/code/}
is employed to implement the BP algorithm.

\subsection{Summary of the Algorithm}
Algorithm 1 illustrates the pipeline of the proposed method.
We first extract raw features from
the PolSAR data (Section III.B).
Then, contextual features are extracted
using the 3D-DWT technique (Section III.C).
Finally, we predict class labels by optimizing our designed
SVM-MRF based supervised learning model (Section III.D-E).

\vspace{-.1cm}
\begin{algorithm}[htb]
\caption{PolSAR Image  {Semantic} Segmentation Using 3D Discrete
Wavelet Transform and Markov Random Field}
\hspace*{0.02in} {\bf Input: }
Pixel-wise coherence matrix $T$, ground truth class labels,
pairwise smoothness parameter $\alpha_s$\\
\hspace*{0.02in} {\bf Output: }
Pixel-wise class labels $\mathbf{Y}$
\begin{algorithmic}[1]
\State Extract raw features $\mathbf{X} \in {R^{H\times{W}\times{D}}}$
\State Convert $\mathbf{X}$ to ${\mathbf{Z}} \in {R^{H\times{W}\times{15D}}}$ using 3D-DWT feature extraction technique
\State Randomly select a proportion of pixels with ground truth class labels as training samples and train a multi-class SVM classifier
\State Estimate class probabilities of pixels using the trained SVM classifier
\State Predict the class labels $\mathbf{Y}$ by solving an MRF model with BP optimization algorithm
\State $\bf end$
\end{algorithmic}
\end{algorithm}
\vspace{-0.6cm}
\begin{figure}[htb]	
	\vspace{-0.5cm}
\begin{center}
		\includegraphics[height=10.0cm,width=8.0cm]{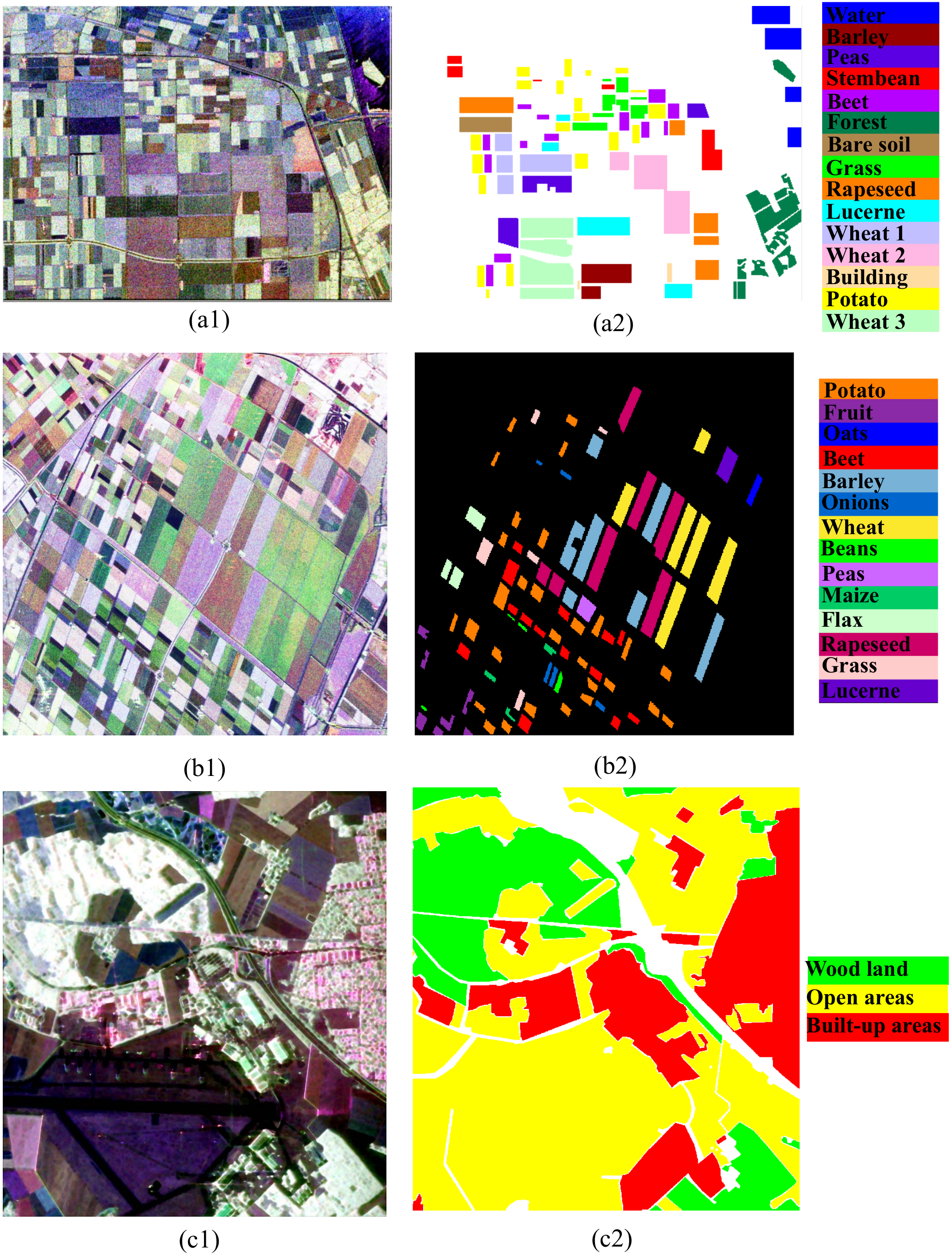}
\vspace{-0.3cm}
		\caption{Experimental images.
(a1) PauliRGB image of Flevoland area data set 1.
(a2) Ground truth class labels and color codes of Flevoland area data set 1.
(b1) PauliRGB image of Flevoland area data set 2.
(b2) Ground truth class labels and color codes of Flevoland area data set 2.
(c1) PauliRGB image of Oberpfaffenhofen area data set.
(c2) Ground truth class labels and color codes of Oberpfaffenhofen area data set.}
\vspace{-0.7cm}
\end{center}
\end{figure}

\section{Experiments}
\subsection{Experimental Data and Settings}
In this section, experiments on {three} real PolSAR data sets
are conducted to validate the performance of the proposed method.
Figure 4 displays the experimental images employed for evaluation.
{The first image was acquired by
NASA/JPL AIRSAR on August 16, 1989.
It is an L-band PolSAR image with size 750$\times$1024.
We denote this data set as Flevoland area data set 1 in following sections.
Figure 4(a1) presents the PauliRGB image of Flevoland area data set 1,
and Fig.~4(a2) shows the ground truth class labels and color codes.
There are 15 classes in total, including
$\emph{water}$, $\emph{barley}$, $\emph{peas}$, $\emph{stembean}$,
$\emph{beet}$, $\emph{forest}$, $\emph{bare soil}$,
$\emph{grass}$, $\emph{rapeseed}$, $\emph{lucerne}$, $\emph{wheat 1}$,
$\emph{wheat 2}$, $\emph{building}$, $\emph{potato}$, and $\emph{wheat 3}$.

\begin{figure*}	[htb]
				\vspace{0cm}
	\begin{center}
				 \includegraphics[height=3.4cm,width=17cm]{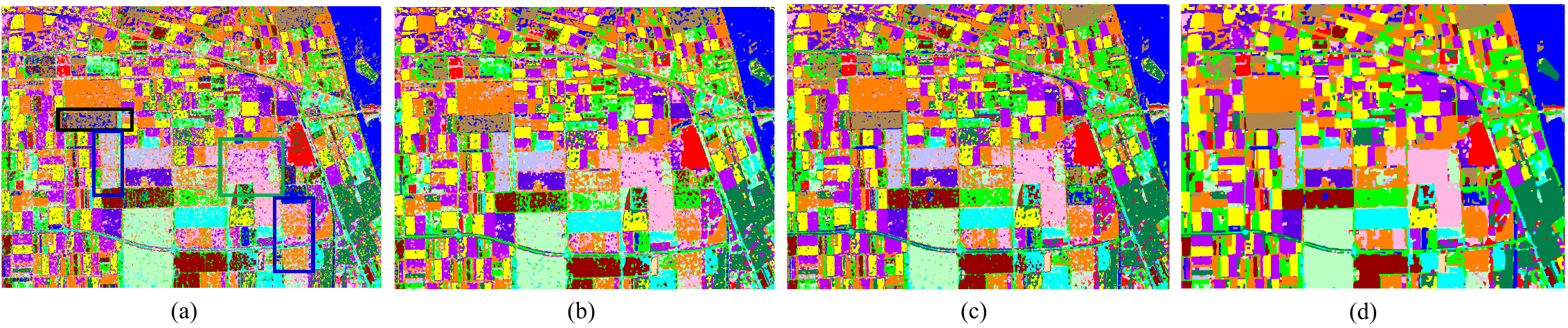}
		\vspace{-.2cm}
		\caption{Ablation study of our proposed method.
			(a) SVM\emph{No}DWT. (b) SVM2D. (c) SVM3D. (d) SVM3D-BPMRF}	
	\end{center}
	\vspace{-.1cm}
\end{figure*}

\vspace{-0cm}
\begin{table*}[htb]
	\vspace{-0.3cm}
	\caption{Ablation study of our proposed method}
	\vspace{-0.5cm}
	\begin{center}
		 \begin{tabular}{c|c|c|c|c|c|c|c|c}
			\hline	\hline
{Method}& \emph{Water}&\emph{Barley}
&\emph{Peas}&\emph{Stembean}&\emph{Beet}&\emph{Forest}&\emph{Bare soil}&\emph{Grass}\\ \hline
{SVM\emph{No}DWT} 		
&{96.12}&{{90.66}}&{88.02} &{80.51}&{73.49}&{83.53}&{74.46}&{45.27}\\
{SVM2D}	
&{94.37}&{95.23}&{92.57} &{95.09}&{87.15}&{90.83}&{92.46}&{59.07}\\
{SVM3D}	
&{97.23}&{95.88}&{92.50}&{95.53} &{94.76}&{93.75}&{96.18}&{67.04}\\
{SVM3D-BPMRF}
&\bf{98.42}&\bf{97.66}&\bf{97.38} &\bf{98.20}&\bf{99.64}&\bf{98.80}&\bf{99.75}&\bf{83.22}
\\ \hline	\hline			
{Method}&\emph{RapeSeed}&\emph{Lucerne}
&\emph{Wheat 1}&\emph{Wheat 2}&\emph{Building}&\emph{Potato}
&\emph{Wheat 3}&{$\emph{Overall CA}$}\\ \hline					
{SVM\emph{No}DWT}
&{72.86}&{76.67}&{64.95}&{82.25}&{45.85}&{81.57}&{88.30}&{79.98}\\
{SVM2D}
&{79.60}&{90.19}&{83.74}&{86.26}&{85.58}&{89.89}&{91.36}&{87.66}\\
{SVM3D}
&{86.37}&{90.20}&{79.23}&{87.97} &{94.56}&{90.90}&{93.21}&{90.57}\\
{SVM3D-BPMRF}
&\bf{95.53}&\bf{95.88}&\bf{89.70}&\bf{96.86}&\bf{99.32}&\bf{97.83}&\bf{98.87}&\bf{96.72}
\\\hline \hline
		\end{tabular}
	\end{center}
\vspace{-0.5cm}
\end{table*}

The second data set is another L-band image
collected by AIRSAR over Flevoland area in 1991.
The size of this image is 1020$\times$1024.
This data set is denoted as Flevoland area data set 2
in Section IV.
Figure 4(b1) displays the PauliRGB image of Flevoland area data set 2.
The ground truth class labels and color codes are presented in Fig.~4(b2).
Flevoland area data set 2 includes 14 classes, which are
$\emph{potato}$, $\emph{fruit}$, $\emph{oats}$, $\emph{beet}$,
$\emph{barley}$, $\emph{onions}$, $\emph{wheat}$,
$\emph{beans}$, $\emph{peas}$, $\emph{maize}$, $\emph{flax}$, $\emph{rapeseed}$,
$\emph{grass}$, and $\emph{lucerne}$, respectively.

Figure 4(c1) shows the PauliRGB image of the third experimental data set.
It is an L-band image obtained by E-SAR, German Aerospace Center,
over Oberpfaffenhofen area in Germany.
The size of this image is 1300$\times$1200.
The ground truth class labels and color codes are given in Fig.~4(c2).
There are three labeled classes in the image: \emph{built-up areas},
\emph{wood land}, and \emph{open areas}.
The void areas are unlabeled class,
which are not taken into consideration during the experiments.}

In the experiments, we first conduct ablation study in
Section III.B, justifying the effectiveness of the two key and novel
components of the proposed method:
3D-DWT contextual features and MRF optimization.
Then parameter analysis will be presented in Section III.C.
Finally, in Section III.D, we demonstrate the
effectiveness of the proposed method
by comparison with the other eight methods listed as below.

(1) {KNN} method \cite{HouBin2016} which employs a Euclidean distance
defined on the basis of three polarimetric PauliRGB channels.

(2) {SupWishart} method \cite{Lee1999} which is supervised method
based on Wishart statistical hypothesis of PolSAR data.

(3) {SupWishart-PMRF} method which combines
supervised Wishart {semantic} segmentation with Potts MRF model.

(4) {MLR\emph{sub}MLL} method \cite{MLRsubMLL} which combines
subspace multinomial logistic regression and MRF.
{Graph cut method is employed to solve the MRF in this method.}

(5) {{GGW-BPMRF} method}.
The original paper \cite{TexfeaMRF2016} employs GLCM, Gabor and wavelet transforms of PauliRGB data as texture features, and
SVM with Potts MRF model as classifier.
{To fairly evaluate the performance of different features,
we change the Potts MRF model to BP MRF model instead.}

(6) {SVM3D-GCMRF} method \cite{CaoXiangyong} which exploits
3D-DWT texture features as input feature of SVM classifier
and feature similarity model for MRF optimization.
Graph cut method is utilized
to solve the MRF model.

(7) CNN method \cite{DeepLearning}
which applies deep neural network in PolSAR image  {semantic} segmentation.

(8) {{CNN-BPMRF}} method
which integrates CNN with BP MRF into a unified framework.

We conducted quantitative comparisons
on the {three} experimental data sets,
wherein the classification accuracies
and time consumptions are reported and analyzed.
For convenience, we define the \textit{classification accuracy (CA)}
of a class as the ratio of the number of pixels
correctly classified for the class to the total number
of pixels in this class.
The \textit{Overall CA} is defined as the ratio
of the number of correctly classified
pixels in the whole image to the total number of
pixels in the image.
For all the experiments, we randomly select 1$\%$ pixels
with ground truth class labels as training samples.
{For all SVM-based methods,
in order to determine the SVM parameters,
we randomly select 200 training samples from the training set to
perform cross-validation.
It should be noted that the cross-validation time consumption
is included in the time cost analysis in Section IV.}
All experiments are implemented on a laptop
with 2.6GHz CPU and 16GB memory.

\subsection{Ablation Study}

To evaluate the effectiveness of the two key components
of the proposed method:
3D-DWT contextual features and MRF optimization,
we conduct four groups of experiments:

(1) Using raw polarimetric features and SVM classification,
without 3D-DWT contextual features and MRF optimization
(Denoted as SVM\emph{No}DWT).

(2) Using 2D-DWT texture features, i.e., conducting DWT only on width and height dimensions but not on feature dimension,
and SVM classification, without MRF optimization (Denoted as SVM2D).

(3) Using 3D-DWT features and SVM classification,
without MRF optimization (Denoted as SVM3D).

{(4)} Using 3D-DWT features
and SVM classification with MRF optimization (Denoted as SVM3D-BPMRF).

Taking {Flevoland area data set 1} for example,
Fig.~5 illustrates the  {semantic} segmentation results
under the above four comparison scenarios.
Figure 5(a) shows the  {semantic} segmentation result of SVM\emph{No}DWT,
where a great deal of misclassified pixels can be observed.
{In the region marked by black rectangle,
an appreciable part of \emph{bare soil} pixels
are misclassified as \emph{water} class.
In addition, we can find distinct class confusions between
\emph{rapeseed} and \emph{wheat 1} [highlighted by blue rectangles],
and \emph{wheat 2} and \emph{beet} [highlighted by green rectangle].}
Except for the above class confusions,
the whole class label map exhibits a speckle-like appearance.
{The reason accounting for these phenomena is that
the representative ability of the raw features is weak,
and the speckle noise is not effectively depressed
without encoding the contextual information.}
Figure 5(b) displays the  {semantic} segmentation
result based on 2D-DWT features.
We can discover from this figure that
the speckle noise is greatly mitigated compared with Fig.~5(a).
The {semantic} segmentation result
using 3D-DWT contextual features is presented in Fig. 5(c).
Compared with the class label map in Fig.~5(b),
less isolated pixels and better spatial consistency
can be observed in this figure,
which exactly illustrates the stronger feature discrimination
and contextual representation of 3D-DWT than 2D-DWT.
The segmentation result of our proposed method is depicted in Fig.~5(d).
It is noteworthy that the segmentation result
after performing MRF optimization is in better conformity
with the ground truth class labels,
and exhibits less intra-class variability \cite{almm2019} and clearer label boundaries,
which demonstrates that MRF optimization can effectively
counteract the speckle noise and promote the contextual consistency.

Tables I lists the numerical results of the four comparison scenarios.
We can conclude from the table that
our proposed method outperforms SVM\emph{No}DWT, SVM2D, and SVM3D
by {16.74\%, 9.06\% and 6.15\%} respectively,
which justifies the effectiveness of feature extraction using 3D-DWT
and the label smoothness enforcement of MRF.
It is noticeable the \emph{Overall CA} value
using 3D-DWT is {2.91\%} higher than 2D-DWT,
which validates the effectiveness of
the wavelet transforms on the third dimension.

\begin{figure}[h]
	\vspace{-.2cm}
	\begin{center}
		\includegraphics[height=5.2cm,width=6.6cm]{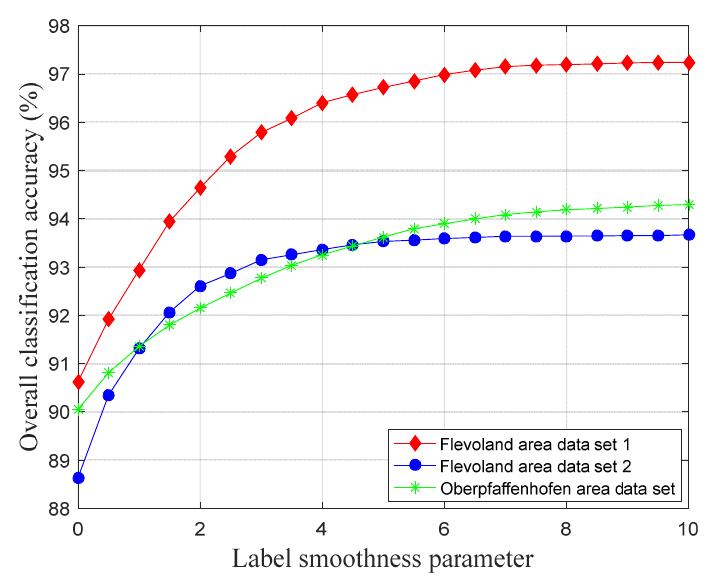}		
		\vspace{-.3cm}
		\caption{The influence of $\alpha_s$ on classification accuracy.}	
	\end{center}
	\vspace{-0.3cm}
\end{figure}

\begin{figure*}[htb]
	\vspace{-.2cm}
	\begin{center}
		\includegraphics[height=3.3cm,width=12.7cm]{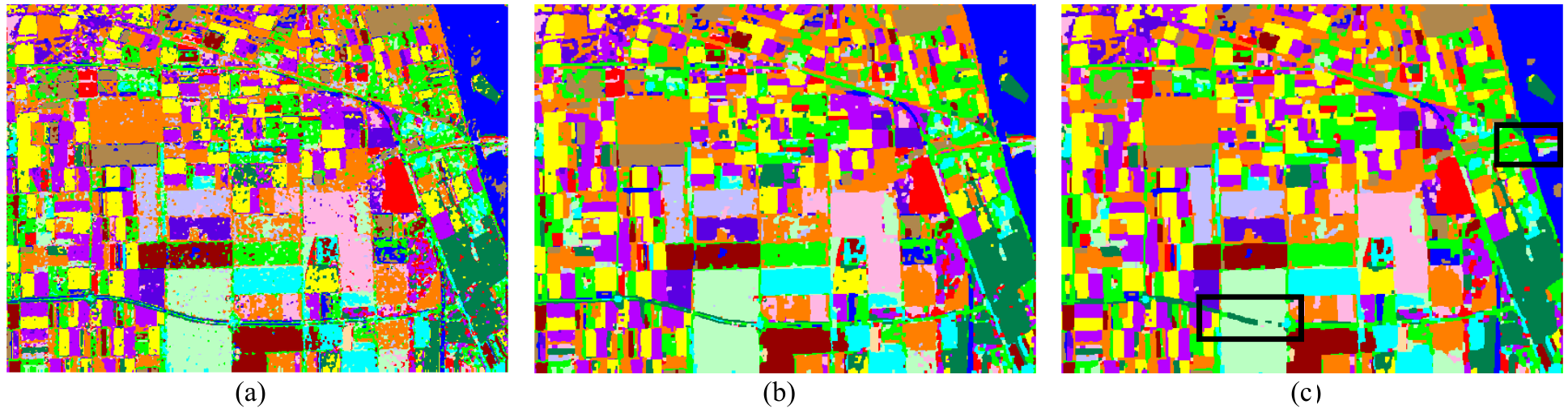}		
		\vspace{-.3cm}
		\caption{ {Semantic} segmentation results for Flevoland area data with different $\alpha_s$. (a) $\alpha_s$=1 (b) $\alpha_s$=5 (b) $\alpha_s$=10.}	
	\end{center}
	\vspace{-.5cm}
\end{figure*}
\vspace{-.5cm}
\subsection{Parameter Analysis}

For the proposed method, parameter $\alpha_s$
which determines the tightness of label
smoothness constraint significantly
influences the final segmentation result.
Larger $\alpha_s$ indicates that label smoothness term
plays a more important role during the optimization,
therefore tends to generate smoother {semantic} segmentation result.
However, too large $\alpha_s$ will lead to loss of details.
Therefore, we next analyze the impact of parameter $\alpha_s$,
to reach a compromise between the
classification accuracy and detail preservation.

Figure 6 shows the influence of parameter $\alpha_s$
on the \emph{Overall CA} value,
where $\alpha_s$ takes value from 0 to 10 with step size 0.5.
$\alpha_s$$=$$0$ corresponds to the scenario without MRF optimization.
It can be observed obviously from Fig.~6 that the classification accuracy
increases significantly with the increment of $\alpha_s$
when $\alpha_s$ is less than 3.
When $\alpha_s$ grows larger than 3,
the increase tendency of \emph{Overall CA} value
slows down and roughly converges at $\alpha_s$$=$$5$.
When $\alpha_s \textgreater 5$, the $\alpha_s$ increase brings
marginal improvement on classification accuracy.

To intuitively illustrate the influence of parameter $\alpha_s$,
taking {Flevoland area data set 1} for example,
we show the  {semantic} segmentation results with $\alpha_s$
taking three typical values 1, 5 and 10 in Fig.~7.
We can observe from Fig.~7(a) that when $\alpha_s$ takes the value 1,
although the whole class label map is smoother than the case
without MRF optimization as shown in Fig.~5(c),
some speckle-like and isolated pixels still scatter in the image.
Moreover, the label boundaries are not clear and straight.
Figure 7(b) exhibits the  {semantic} segmentation result with $\alpha_s=5$.
We can discover from the figure that
the speckle noise is better depressed than the result shown in Fig. 7(a),
and the label boundaries are well aligned with the image edges.
However, when $\alpha_s$ increases to value 10,
although the whole image shows favorable spatial connectivity,
many terrain details fade out or even disappear in the segmentation map,
especially the thin and small regions.
For example, the bridge in the right middle part
and the road in the middle bottom part of the image
[marked by black rectangles]
get broken after strong smoothing.
Therefore, to make a balance between the spatial connectivity and
detail preservation, $\alpha_s$ takes the value 5 in the following experiments.

\begin{figure*}[htb]
				\vspace{-.1cm}
	\begin{center}				 \includegraphics[height=10.3cm,width=12.5cm]{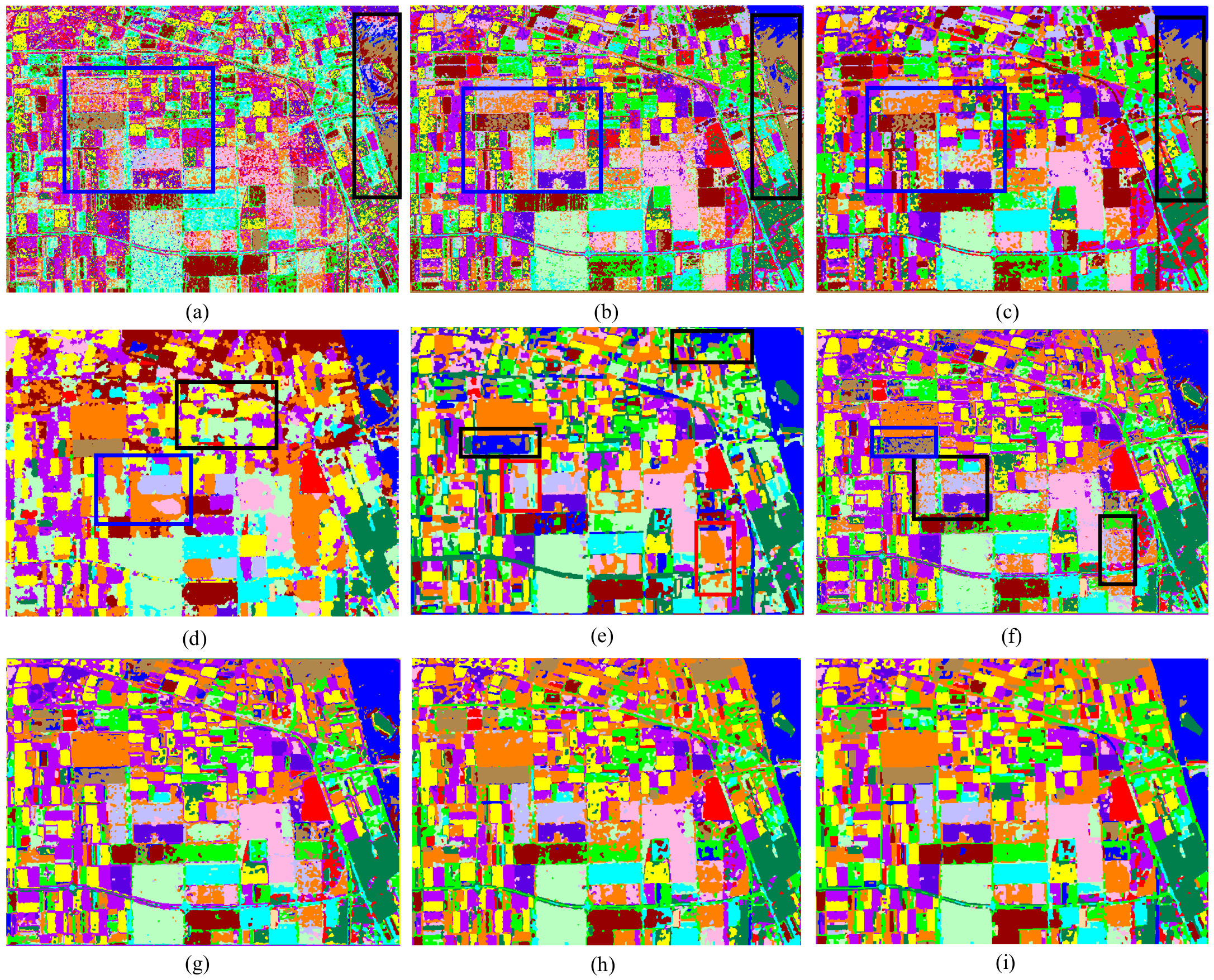}		
		\vspace{-.3cm}
		\caption{ {Semantic} segmentation results of different methods on Flevoland area data set 1. (a) KNN. (b) SupWishart. (c) SupWishart-PMRF.
			(d) MLR\emph{sub}MLL. (e) GGW-BPMRF. (f) CNN.
			(g) CNN-BPMRF. (h) SVM3D-GCMRF.  (i) SVM3D-BPMRF.}	
	\end{center}
	\vspace{-.5cm}
\end{figure*}

\begin{table*}[htb]
				\vspace{-.5cm}
	\caption{CAs (\%) of Flevoland area data set 1 with different methods}
	\vspace{-.5cm}
	\begin{center}
		\begin{tabular}{c|c|c|c|c|c|c|c|c}
			\hline	\hline
			{Method}& \emph{Water}&\emph{Barley}
			&\emph{Peas}&\emph{Stembean}&\emph{Beet}&\emph{Forest}&\emph{Bare soil}&\emph{Grass}\\ \hline
			{KNN \cite{HouBin2016}} 	
			&{18.84}&{88.73}&{62.00} &{25.24}&{70.40}&{41.25}&{81.31}&{38.00}\\
			{SupWishart \cite{Lee1999}}
			&{52.61}&{45.77}&{88.33} &{83.01}&{74.98}&{86.42}&{59.29}&{65.09}\\
			{SupWishartMRF}
			&{54.84}&{50.70}&{95.69}&{92.41} &{76.95}&{92.39}&{56.43}&{71.55}\\
			{MLR\emph{sub}MLL \cite{MLRsubMLL}}
			&{91.94}&{81.54}&{0.02} &{84.79}&{80.50}&{96.25}&{98.10}&{0.00}\\
			{GGW-BPMRF}
			 &{92.89}&{92.35}&{88.71}&{78.49}&{90.22}&{96.80}&{62.93}&{56.15}\\
			{CNN \cite{DeepLearning}}
			&{95.38}&{92.48}&{96.19} &{97.02}&{96.11}&{97.95}&{79.62}&{77.50}\\
			{CNN-BPMRF}
			 &{97.42}&{96.17}&\bf{98.28}&{97.78}&{98.60}&{98.65}&{91.64}&\bf{83.51}\\
			{SVM3D-GCMRF \cite{CaoXiangyong}}
			&{96.37}&{97.35}&{95.88}&{96.88}&{95.76}&{97.66}&\bf{100.00}& {82.01}\\
			{Our Method}
			 &\bf{98.42}&\bf{97.66}&{{97.38}}&\bf{98.20}&\bf{99.64}&\bf{98.80}&{99.75}&
			{83.22}\\
			\hline	\hline			
			\bf{Time Cost}& \emph{RapeSeed}&\emph{Lucerne}
			&\emph{Wheat 1}&\emph{Wheat 2}&\emph{Building}&\emph{Potato}
			&\emph{Wheat 3}&\bf{$\emph{Overall CA}$}\\ \hline					
			{5.8 s}
			 &{35.98}&{46.03}&{46.97}&{56.20}&{65.71}&{49.01}&{65.79}&{50.84}\\
			{10.2 s}
			 &{46.96}&{86.77}&{14.71}&{92.38}&{84.08}&{68.67}&{83.89}&{70.00}\\
			{19.2 s}
			&{59.37}&{84.92}&{13.15}&\bf{97.25} &{80.95}&{65.16}&{87.16}&{73.49}\\
			{6 h 28 m 21.8 s}
			 &{90.87}&{91.93}&{38.42}&{64.08}&{86.39}&{91.09}&{79.47}&{73.91}\\
			{4 m 10.9 s}
			 &{74.57}&{69.83}&{47.36}&{92.61}&{60.68}&{75.84}&{95.53}&{82.48}\\
			{35 m 26.7 s}
			 &{76.16}&{93.57}&{78.53}&{87.46}&{79.86}&{90.26}&{94.57}&{90.18}\\
			{37 m 15.1 s}
			 &{81.06}&{94.62}&{88.47}&{92.11}&{84.49}&{94.10}&{97.22}&{93.83}\\
			{6 h 30 m 52.3 s}
			 &{91.03}&{94.66}&\bf{91.61}&{96.23}&{77.55}&{94.70}&{97.49}&{95.04}\\
			{5 m 6.0 s}
			 &\bf{95.53}&\bf{95.88}&{89.70}&{96.86}&\bf{99.32}&\bf{97.83}&\bf{98.87}&\bf{96.72}
			\\\hline \hline
		\end{tabular}
	\end{center}
	\vspace{-.5cm}
\end{table*}

\begin{figure*}[htb]
	\vspace{0cm}
	\begin{center}		 \includegraphics[height=11cm,width=10.6cm]{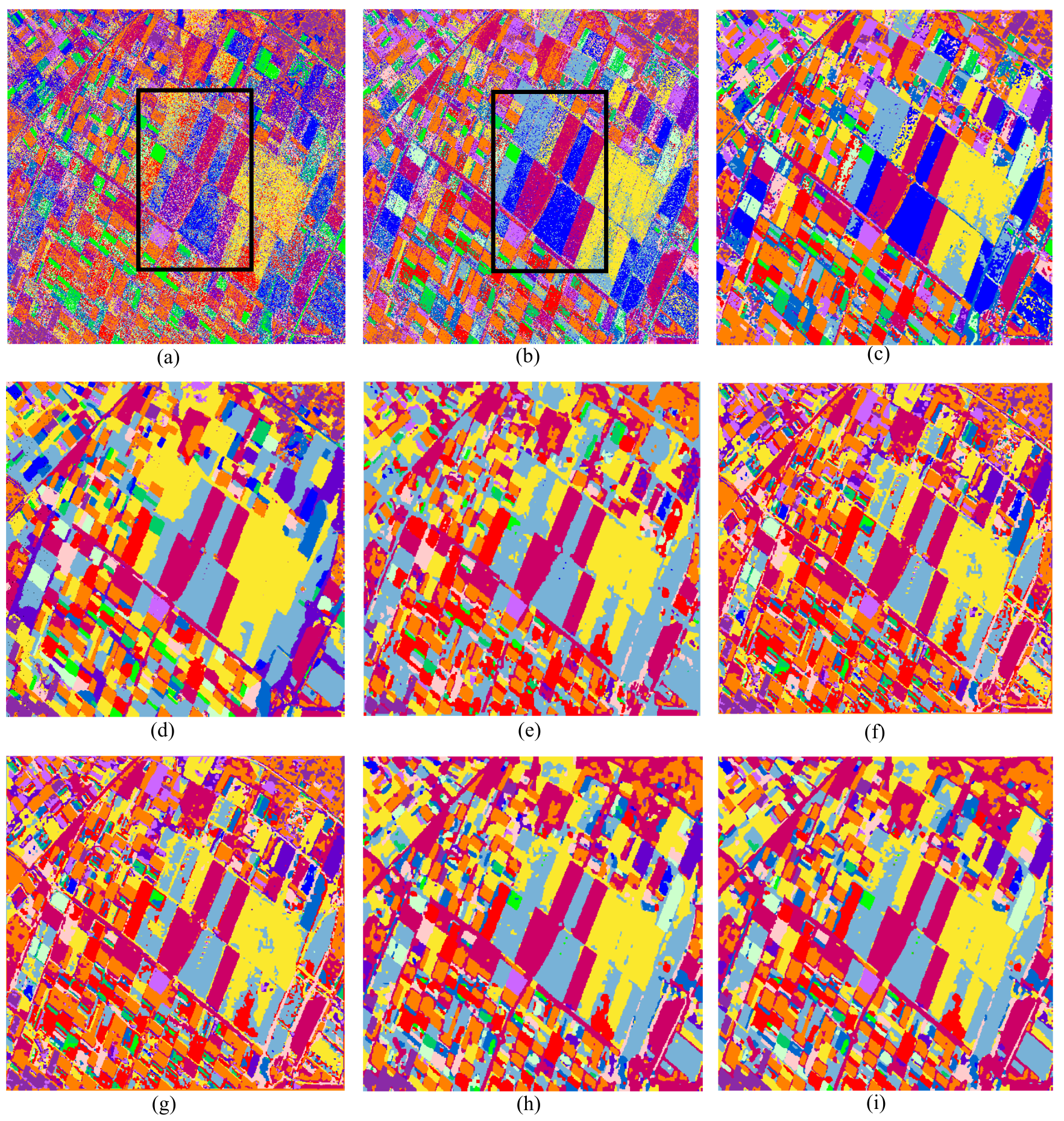}
		\vspace{-.4cm}		
		\caption{ {Semantic} segmentation results of different methods on Flevoland area data set 2. (a) KNN. (b) SupWishart. (c) SupWishart-PMRF.
			(d) MLR\emph{sub}MLL. (e) GGW-BPMRF. (f) CNN.
			(g) CNN-BPMRF. (h) SVM3D-GCMRF. (i) SVM3D-BPMRF.}	
	\end{center}
	\vspace{-.5cm}
\end{figure*}

\begin{table*}[htb]
				\vspace{-.5cm}
	\caption{CAs (\%) of Flevoland area data set 2 with different methods}
	\vspace{-.5cm}
	\begin{center}
		\begin{tabular}{c|c|c|c|c|c|c|c|c}
			\hline	\hline
			{Method}&\emph{Potato}&\emph{Fruit}
			 &\emph{Oats}&\emph{Beet}&\emph{Barley}&\emph{Onions}&\emph{Wheat}&\bf{\emph{Overall CA}}\\ \hline
			{KNN \cite{HouBin2016}} 	
			&{67.92}&{80.97}&{81.64} &{39.93}&{30.99}&{18.54}&{53.68}&{48.89}\\
			{SupWishart \cite{Lee1999}}
			&{87.24}&{26.33}&{88.52} &{67.93}&{32.39}&{17.18}&{77.85}&{65.78}\\
			{SupWishartMRF}
			&{98.46}&{13.37}&{94.98}&{83.99} &{27.32}&{11.92}&{89.89}&{73.06}\\
			{MLR\emph{sub}MLL \cite{MLRsubMLL}}
			&{83.15}&{98.24}&\bf{99.71} &{51.07}&{74.79}&{32.08}&{79.95}&{79.58}\\
			{GGW-BPMRF}
			 &{91.66}&{92.76}&{81.64}&{82.33}&{95.73}&{4.51}&{93.13}&{86.83}\\
			{CNN \cite{DeepLearning}}
			&{90.18}&{93.82}&{68.36} &{72.37}&{83.09}&{32.86}&{88.49}&{86.16}\\
			{CNN-BPMRF}
			 &{92.81}&{96.37}&{83.00}&{76.10}&{87.50}&{33.05}&{93.44}&{89.75}\\
			{SVM3D-GCMRF \cite{CaoXiangyong}}
			&\bf{95.54}&{99.90}&{79.34}&\bf{79.01}&{97.53}&{55.49}&{98.20}& {93.01}\\
			{Our Method}
			 &{95.01}&\bf{99.91}&{79.25}&{78.48}&\bf{98.44}&\bf{57.79}&\bf{98.77}&
			\bf{93.43}\\
			\hline	\hline			
			{Method}&\emph{Beans}& {Peas}&\emph{Maize}&\emph{Flax}
			&\emph{Rapeseed}&\emph{Grass}&\emph{Lucerne}&\bf{Time Cost}\\ \hline				 
			{KNN \cite{HouBin2016}}
			&{60.63}&{78.38}&{48.60}&{32.95}&{45.88}&{31.83}&{58.27}&{7.3 s}\\
			{SupWishart \cite{Lee1999}}
			&{38.17}&{89.54}&{60.85}&{85.98}&{74.59}&{38.18}&{74.09}&{13.1 s}\\
			{SupWishartMRF}
			&{32.16}&\bf{100.00}&{57.75}&\bf{94.56} &{85.67}&{46.91}&{86.65}&{25.9 s}\\
			{MLR\emph{sub}MLL \cite{MLRsubMLL}}
			&\bf{79.92}&{98.70}&{45.99}&{91.24}&{92.07}&{56.98}&{88.50}&{12 h 2 m 52.2 s}\\
			{GGW-BPMRF}
			&{31.15}&{99.81}&\bf{82.87}&{18.09}&{94.33}&{76.64}&{46.54}&{5 m 11.5 s}\\
			{CNN \cite{DeepLearning}}
			&{53.97}&{98.47}&{61.09}&{90.17}&{96.55}&{66.06}&{84.69}&{41 m 8.4 s}\\
			{CNN-BPMRF}
			&{54.90}&{99.21}&{82.17}&{94.19}&\bf{97.92}&{71.00}&{88.96}&{43 m 17.6 s}\\
			{SVM3D-GCMRF \cite{CaoXiangyong}}
			&{62.57}&{99.68}&{77.36}&{91.26}&{93.53}&{81.66}&{92.04}&{12 h 9 m 38.2 s}\\
			{Our Method}
			 &{62.88}&\bf{100.00}&{72.84}&{92.92}&{94.09}&\bf{82.24}&\bf{95.67}&{7 m 22.9 s}\\
			\hline \hline
		\end{tabular}
	\end{center}
	\vspace{-.5cm}
\end{table*}

\begin{figure*}[htb]
	\vspace{0cm}
	\begin{center}		 \includegraphics[height=11cm,width=10.8cm]{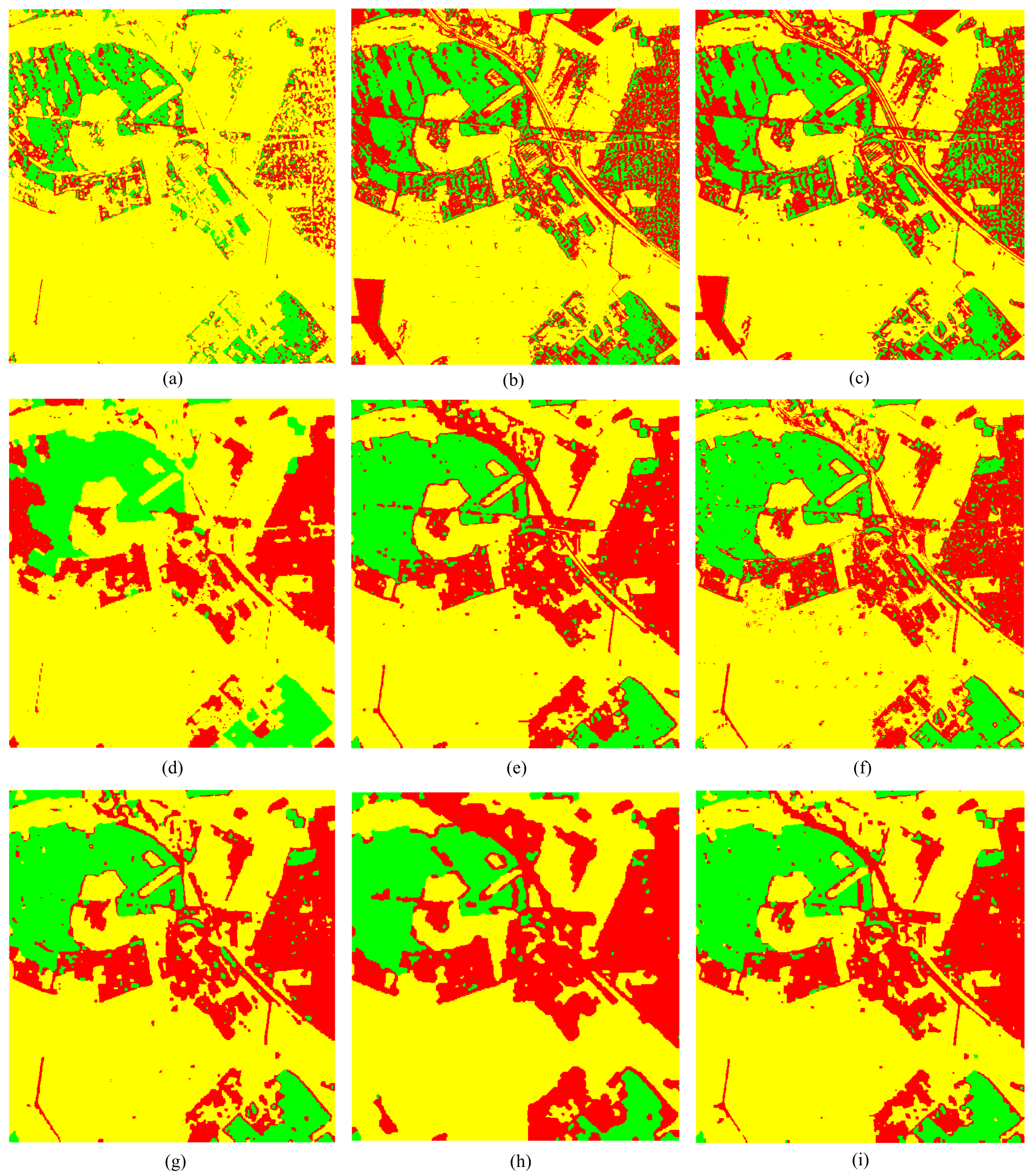}
		\vspace{-.3cm}		
		\caption{ {Semantic} segmentation results for Oberpfaffenhofen area data set for different methods. (a) KNN. (b) SupWishart. (c) SupWishart-PMRF.
			(d) MLR\emph{sub}MLL. (e) GGW-BPMRF. (f) CNN.
			(g) CNN-BPMRF. (h) SVM3D-GCMRF. (i) SVM3D-BPMRF.}	
	\end{center}
	\vspace{-.6cm}
\end{figure*}

\begin{table*}[htb]
	\vspace{-.5cm}
	\begin{center}
		\caption{CAs (\%) of Oberpfaffenhofen area data set with different methods}
			\vspace{-.2cm}
		\begin{tabular} {c|c|c|c|c|c} \hline \hline
			{Method}&\emph{Built-up areas}&\emph{Wood land}&\emph{Open areas} &\bf{\emph{Overall CA}}&\bf{{Time Cost}}\\ \hline
			{KNN \cite{HouBin2016}}           & {57.83} &{23.55} &\bf{99.75} &{72.95}&{1 m 25.0 s}  \\
			{SupWishart \cite{Lee1999}}       & {66.60} &{56.28} &{90.66} &{77.55} &{10.9 s}\\
			{SupWishart-PMRF}        & {70.27} &{55.72} &{90.04} &{77.79} &{33.4 s}\\
			{MLR\emph{sub}MLL \cite{MLRsubMLL}} & {74.52} &{72.28} &{99.68} &{88.07}&{27 h 5 m 51.2 s}\\
			{GGW-BPMRF} & {89.25} &\bf{87.38} &{97.26} &{93.28} &{12 m 25.3 s}  \\
			{CNN \cite{DeepLearning}}	       & {89.26} &{78.95} &{97.23} &{91.21} &{54 m 32.6 s}\\
			{CNN-BPMRF}	       & \bf{90.91} &{84.43} &{98.02} &{93.51}&{56 m 18.0 s}\\
			{SVM3D-GCMRF \cite{CaoXiangyong}}  & {80.11} &{92.09} &{93.89} &{90.73}&{27 h 18 m 45.7 s }  \\
			{Our method}           & {90.40} &{86.68} &{97.77} &\bf{93.62}&{15 m 31.9 s}  \\
			\hline \hline
		\end{tabular}
	\end{center}
	\vspace{-.3cm}
\end{table*}

\vspace{-4mm}
\subsection{Results and Comparisons}
\subsubsection{Flevoland Area Data set 1}
The class label maps of Flevoland area data
using the nine compared methods are presented in Fig.~8,
and the \emph{CA} values and time consumptions are reported in Table II.

Figure 8(a)-(b) illustrate the class label maps
using KNN and SupWishart methods respectively.
We can find a great number of miscellaneous labels
in these two figures [marked by the black and blue rectangles in Fig. 8(a)-(b)].
This is because the KNN and Wishart distances are not discriminative enough
to distinguish the confused classes with similar scattering mechanisms.
SupWishart-PMRF method integrates Potts MRF priors with SupWishart method,
achieving better spatial connectivity.
However, the confusion between {\emph{water} and \emph{bare soil},
\emph{rapeseed} and \emph{wheat 1}} classes
can still be noticeably observed
[marked by the black and blue rectangles in Fig. 8(c)].
Figure 8(d) presents the class label map of MLR\emph{sub}MLL method.
We can find obvious ambiguities
between {\emph{rapeseed} and \emph{wheat 3}} classes,
and {\emph{rapeseed} and \emph{wheat 1}} classes
[marked by the blue rectangle in Fig.~7(d)].
{In addition, MLR\emph{sub}MLL method fails to discriminate
\emph{grass} class [highlighted by black rectangle in Fig.~7(d)].}
The numerical results in Table II
show that the $\emph{Overall CA}$ values of
KNN, SupWishart, SupWishart-PMRF, and MLR\emph{sub}MLL
lag behind our proposed method by {45.88\%, 26.72\%, 23.23\%
and 22.81\%} respectively.

To better validate the effectiveness of our proposed method,
we compare it with four state-of-the-art methods,
i.e., GGW-BPMRF, SVM3D-GCMRF, CNN and CNN-BPMRF methods.
{Investigating Fig.~8(e)-(i) and Table II, we can conclude:

(1) 3D-DWT features are more representative than combinations of
2D texture features. Using the same BP MRF model,
the superiority of the $\emph{Overall CA}$ value of
our proposed method over GGW-BPMRF
is owing to the stronger ability of 3D-DWT features in
distinguishing diverse terrains.

(2) CNN fails to differentiate a number of classes
with similar scattering mechanisms
[marked by black and blue rectangles in Fig.~8(f)].
This is because only 1,677 training samples
(1\% pixels with ground truth class labels) are available
in this experimental set,
which results in insufficient tuning of the CNN filters
due to the deep architecture of the network.
The $\emph{Overall CA}$ values of CNN and CNN-BPMRF methods
fall behind our proposed method by
6.54\% and 2.89\% respectively.
Besides, our proposed method consumes much less time (5 minutes 6.0 seconds)
than CNN (35 minutes 26.7 seconds) and CNN-BPMRF (37 minutes 15.1 seconds).

(3) With the same 3D-DWT features,
our proposed method marginally outperforms SVM3D-GCMRF method by 1.68\%.
However, SVM3D-GCMRF method consumes 6 hours 30 minutes 52.3 seconds
on the model convergence, while our proposed method
only costs 5 minutes 6.0 seconds.
The distinct lower time consumption makes
our proposed method more practical in real applications.

The above conclusions gracefully demonstrate
the strong discriminative ability of 3D-DWT features
under low annotation scenario, and the favorable label
smoothness enforcement of MRF optimization
with low time consumption.}

{
\subsubsection{Flevoland Area Data set 2}
Figure 9 illustrates the class label maps of the nine compared methods
on Flevoland area data set 2.
The numerical results of their \emph{CA} values and time consumptions
are reported in Table III.
Observing Fig.~9(a)-(i) and Table III,
we highlight below main observations:

(1) Without utilizing texture features or MRF optimization,
the speckle noise in Fig.8~(a)-(b)
greatly degrades the segmentation performance.
The erroneous classifications of KNN and SupWishart methods are severe,
where \emph{oats} and \emph{barley} classes are totally confused
[marked by black rectangles].
In addition, the two images present granular appearances,
which is attributed to the impact of speckle noise.

(2) The performance of CNN is undesirable
when the annotated data are scarce.
We can discover from Table III that
the \emph{CA} values of CNN
on \emph{onions}, \emph{beans}, \emph{maize} and \emph{grass} classes
are 32.86\%, 53.97\%, 61.09\%, and 66.06\% respectively,
which far trails behind  the performance of our proposed method.
This is because the training samples of these four classes
are too limited for CNN to adequately train the deep-structured network.

(3) MLR\emph{sub}MLL and SVM3D-GCMRF methods
which employ graph cut algorithm to optimize the MRF model,
consume far more time
(12 hours 2 minutes 52.2 seconds for MLR\emph{sub}MLL
and 12 hours 9 minutes 38.2 seconds for SVM3D-GCMRF, respectively)
than methods using BP algorithm (5 minutes 11.5 seconds for GGW-BPMRF,
and 7 minutes 22.9 seconds for our proposed method).

(4) Our proposed method again yields the best \emph{Overall CA} value
among the nine competitors.
The combination of 3D-DWT features and MRF optimization
significantly alleviates the negative impact of speckle noise,
achieving preferable segmentation performance and contextual consistency.}

{
\subsubsection{Oberpfaffenhofen Area Data set}
The semantic segmentation results of different methods
on Oberpfaffenhofen area data set
are presented in Fig.~10, and the numerical results
of \emph{CA} values and time consumptions are listed in Table IV.

The Oberpfaffenhofen area data set has two characteristics.
Firstly, its size (1300 $\times$ 1200) is the largest
among the three experimental data sets.
Secondly, the training set size of this data set is 13,721,
which is much larger than Flevoland area data set 1 (1,677 training samples) and Flevoland area data set 2 (1,353 training samples).
The above two characteristics make Oberpfaffenhofen area data set
a favorable experimental scenario
for analyzing the time consumptions of different methods.
From Fig.~10 and Table IV, we can conclude:

(1) The segmentation performance and spatial connectivity
of the methods without texture features, i.e.,
KNN, SupWishart and SupWishart-PMRF methods,
are obviously inferior to the remaining methods,
which demonstrates the effectiveness of texture features
in PolSAR image segmentation.

(2) The performance of GGW-BPMRF, CNN-BPMRF and our proposed method
are very close to each other.
Benefitting from the contextual information incorporation
using both texture features and MRF optimization,
the class label maps of these methods present a desirable visual effect
with preferable contextual consistency and clear label boundaries
while well preserving the image details.
It should be highlighted that, due to the complex learning process,
CNN-BPMRF method consumes more time (56 minutes 18.0 seconds)
than our proposed method (15 minutes 31.9 seconds).

(3) Investigating Fig.10~(d) and (h),
we can discover that the graph cut method
employed in MLR\emph{sub}MLL and SVM3D-GCMRF methods
over-smooths the class label maps,
where a great number of terrain details,
especially some thin and small regions,
fade out or even disappear in the segmentation results.
This may be due to the fact that compared with
Flevoland area data set 1 and Flevoland area data set 2,
the image boundaries in the Oberpfaffenhofen area data set are
more complex and oblique.
It is noteworthy that the time consumptions of these two methods
are much longer (27 hours 5 minutes 51.2 seconds for MLR\emph{sub}MLL,
and 27 hours 18 minutes 45.7 seconds for SVM3D-GCMRF)
than the methods using BP algorithm.}

\vspace{-0.3cm}
\section{Conclusion}
In this paper, we presented a PolSAR image {semantic} segmentation method with 3D discrete wavelet transform (3D-DWT) and Markov random field (MRF).
{The advantages of our work lie in three points:
(1) The proposed method incorporates contextual {semantic} information
in both the feature extraction and post-optimization process,
which effectively depresses speckle noise and enforces spatial consistency.
(2) 3D-DWT is an effective polarimetric discriminator which
can differentiate diverse terrains under low annotation scenario.
(3) Our defined MRF effectively enforces class label smoothness
and the alignment of label boundaries with the image edges.
The employed belief propagation (BP) optimization algorithm
solves our defined MRF model with high efficiency.

To further ameliorate the issues caused
by low annotations and speckle noise,
we plan to develop below approaches in the future:
(1) Feature extraction method based on subspace image factorization,
which enables effective feature extraction
while removing noise simultaneously.
(2) Deep learning based multi-level feature fusion
which enforces two directional, i.e., top-bottom and bottom-top,
feature fusions to suppress noise.

\ifCLASSOPTIONcaptionsoff
  \newpage
\fi




\begin{thebibliography}{1}
\bibitem{HeWL2013}
C.~He, S.~Li, Z.~X. Liao, and M.~S. Liao,
``Texture classification of PolSAR data based on sparse coding of wavelet polarization textons,''
\emph{IEEE Trans. Geosci. Remote Sens.}, vol. 51, no. 8, pp. 4576-4590, Aug. 2013.


\bibitem{TexfeaMRF2016}
A.~Masjedi, M.~J. V, Zoej, and Y.~Maghsoudi,
``Classification of polarimetric SAR images based on modeling contextual information and using texture features,''
\emph{IEEE Trans. Geosci. Remote Sens.}, vol. 54, no. 2, pp. 932-943, Feb. 2016.


\bibitem{DeepLearning}
Y.~Zhou, H.~Wang, F.~Xu, and Y.~Jin,
``Polarimetric SAR image classification using deep convolutional neural networks,''
\emph{IEEE Geosci. Remote Sens. Lett.}, vol. 13, no. 12, pp. 1935-1939, Dec. 2016.


\bibitem{TGRS_Bi2}
H.~Bi, J.~Sun, and Z.~Xu,
``A Graph-Based Semisupervised Deep Learning
Model for PolSAR Image Classification,''
\emph{IEEE Trans. Geosci. Remote Sens.}, vol. 57, no. 4, pp. 2116-2132, Apr. 2019.


\bibitem{TGRS_Bi3}
H.~Bi, F.~Xu, Z.~Wei, Y.~Xue and Z.~Xu,
``An Active Deep Learning Approach for minimally-Supervised PolSAR Image Classification,''
\emph{IEEE Trans. Geosci. Remote Sens.}, vol. 57, no. 11, pp. 9378-9395, Nov. 2019.

\bibitem{rasti2020}
B.~Rasti, D.~Hong, R.~Hang, P.~Ghamisi, X.~Kang, J.~Chanussot, and J.~Benediktsson, ``Feature extraction for hyperspectral imagery: The evolution from shallow to deep,'' \emph{IEEE Geosci. Remote Sens. Mag.}, 2020. DOI: 10.1109/MGRS.2020.2979764.

\bibitem{HouBin2016}
B. Hou, H. Kou, and L. Jiao,
``Classification of polarimetric SAR images using multilayer autoencoders and superpixels ,''
\emph{IEEE J. Sel. Topics Appl. Earth Observ. Remote Sens.}, vol. 9, no. 7, pp. 3072-3081, Jul. 2016.


\bibitem{WuMRF2008}
Y.~H. Wu, K.~F. Ji, W.~X. Yu, and Y.~Su,
``Region-based classification of polarimetric SAR images using Wishart MRF,''
\emph{IEEE Geosci. Remote Sens. Lett.}, vol. 5, no. 4, pp. 668-672, Oct. 2008.



\bibitem{Doulgeris2015}
A.~P. Doulgeris,
``An automatic $\mathcal{U}$-distribution and Markov Random Field segmentation algorithm for PolSAR images,''
\emph{IEEE Trans. Geosci. Remote Sens.}, vol. 53, no. 4, pp. 1819-1827, Apr. 2015.


\bibitem{TGRS_Bi}
H.~Bi, J.~Sun, and Z.~Xu,
``Unsupervised PolSAR Image Classification Using Discriminative Clustering,''
\emph{IEEE Trans. Geosci. Remote Sens.}, vol. 55, no. 6, pp. 3531-3544, Jun. 2017.



\bibitem{hong2019}
D.~Hong, N.~Yokoya, J.~Chanussot, J.~Xu, and X.~Zhu, ``Learning to propagate labels on graphs: An iterative multitask regression frameworkfor semi-supervised hyperspectral dimensionality reduction,'' \emph{ISPRS J.Photogramm. Remote Sens.}, vol. 158, pp. 35-49, 2019.


\bibitem{yao2019}
J.~Yao, D.~Meng, Q.~Zhao, W.~Cao, and Z.~Xu, ``Nonconvex-sparsity and nonlocal-smoothness-based blind hyperspectral unmixing,'' \emph{IEEE Trans. Image Process.}, vol. 28, no. 6, pp. 2991-3006, 2019.


\bibitem{hong2020}
D.~Hong, X.~Wu, P.~Ghamisi, J.~Chanussot, N.~Yokoya, and X.~Zhu, ``Invariant attribute profiles: A spatial-frequency joint feature extractorfor hyperspectral image classification,'' \emph{IEEE Trans. Geosci. Remote Sens.}, 2020. DOI:10.1109/TGRS.2019.2957251.

\bibitem{Cloude1997}
S.~R. Cloude and E.~Pottier,
``An entropy based classification scheme
for land application of polarimetric SAR,''
\emph{IEEE Trans. Geosci. Remote Sens.}, vol. 35, no. 1, pp. 68-78, Jan. 1997.


\bibitem{Freeman1998}
A.~Freeman and S.~L. Durden,
``A Three-component scattering model for polarmetric SAR data,''
\emph{IEEE Trans. Geosci. Remote Sens.}, vol. 36, no. 3, pp. 963-973, May 1998.


\bibitem{Fourcom2005}
Y. Yamaguchi, T.~Moriyama, M.~Ishido, and H.~Yamada,
``Four-component scattering model for polarimetric SAR image decomposition,''
\emph{IEEE Trans. Geosci. Remote Sens.}, vol. 43, no. 8, pp. 1699-1706, Aug. 2005.


\bibitem{Wishart1994}
J.~S. Lee, K.~W. Hoppel, S.~A. Mango, and A.~R. Miller,
``Intensity and
phase statistics of multilook polarimetric and interferometric SAR imagery,''
\emph{IEEE Trans. Geosci. Remote Sens.}, vol. 32, no. 5, pp. 1017-1028, Sep. 1994.


\bibitem{Lee1994}
J.~S. Lee, M.~R. Grunes, and R.~Kwok,
``Classification of multi-look
polarimetric SAR imagery based on complex Wishart distribution,''
\emph{Int. J. Remote Sens.}, vol. 15, no. 11, pp. 2299-2311, Jul. 1994.


\bibitem{Lee1999}
J.~S. Lee, M.~R. Grunes, T.~L. Ainsworth, L.~J. Du, D.~L. Schuler, and S.~R. Cloude,
``Unsupervised classification using polarimetric decomposition and the complex Wishart classifier,''
\emph{IEEE Trans. Geosci. Remote Sens.}, vol. 37, no. 5, pp. 2249-2258, Sep. 1999.


\bibitem{Lee_Kdistri1994}
J.~S. Lee, D.~L. Schuler, R.~H. Langz, and K.~J. Ransod,
``$K$-distribution for multi-look processed poarimetric SAR imagery,''
In \emph{Proc. IEEE IGARSS}, 1994, vol. 4, pp. 2179-2181.


\bibitem{Doulgeris2011}
A.~P. Doulgeris, S.~N. Anfinsen, and T. Eltoft,
``Automated non-Gaussian
clustering of polarimetric synthetic aperture radar images,''
\emph{IEEE Trans. Geosci. Remote Sens.}, vol. 49, no. 10, pp. 3665-3676, Oct. 2011.


\bibitem{Kong1988}
J.~A. Kong, A.~A. Swartz, H.~A. Yueh, L.~M. Novak, and R.~T. Shin,
``Identification of terrain cover using the optimum polarimetric classifier,''
\emph{J. Electromagn. Waves Applic.}, vol. 2, no. 2, pp. 171-194, 1988.


\bibitem{Pottier1991}
E.~Pottier and J.~Saillard,
``On radar polarization target decomposition theorems with application
to target classification by using network method,''
In \emph{Proc. ICAP}, Apr. 1991, vol. 1, pp. 265-268.


\bibitem{Antropov2014}
O.~Antropov, R.~Rauste, H.~Astola, J.~Praks, T.~Hame, and M.~T. Hallikainen,
``Land cover and soil type mapping from spaceborne PolSAR data at L-Band with probabilistic neural network,''
\emph{IEEE Trans. Geosci. Remote Sens.}, vol. 52, no. 9, pp. 5256-5270, Sep. 1998.


\bibitem{cospace2019}
D.~Hong, N.~Yokoya, J.~Chanussot, and X.~Zhu, ``CoSpace:  Common subspace learning from hyperspectral-multispectral correspondences,'' \emph{IEEE Trans. Geosci. Remote Sens.}, vol. 57, no. 7, pp. 4349-4359, 2019.


\bibitem{Fukuda2001}
S.~Fukuda and H.~Hirosawa,
``Support vector machine classification of land cover: Application to polarimetric SAR data,''
in \emph{Proc. IEEE IGARSS}, Jul. 2001, vol. 1, pp. 187-189.


\bibitem{WishartDBN}
F. Liu, L. Jiao, B. Hou, and S. Yang,
``POL-SAR image classification based on Wishart DBN and local spatial information,''
\emph{IEEE Trans. Geosci. Remote Sens.}, vol. 54, no. 6, pp. 3292-3308, Jun. 2016.


\bibitem{SSA2016}
L. Zhang, W. Ma, and D. Zhang,
``Stacked sparse autoencoder in PolSAR data classification using local spatial information,''
\emph{IEEE Geosci. Remote Sens. Lett.}, vol. 13, no. 9, pp. 1359-1363, Sep. 2016.


\bibitem{WishartDSN}
L. Jiao, and F. Liu,
``Wishart deep stacking network for fast POLSAR image classification,''
\emph{IEEE Trans. Image Process.}, vol. 25, no. 7, pp. 3273-3286, Jul. 2016.


\bibitem{SAESDPL}
Y.~Chen, L.~Jiao, Y.~Li, L.~Li, D.~Zhang, B.~Ren, and N.~Marturi,
``A novel semicoupled projective dictionary pair learning method for PolSAR image classification,''
\emph{IEEE Trans. Geosci. Remote Sens.}, vol. 57, no. 4, pp. 2407-2418, Apr. 2019.


\bibitem{TMF2014}
G.~Liu, M.~Li, Y.~Wu, P.~Zhang, L.~Jia, and H.~Liu,
``PolSAR image classification based on Wishart TMF with specific auxiliary field,''
\emph{IEEE Geosci. Remote Sens. Lett.}, vol. 11, no. 7, pp. 1230-1234, Jul. 2014.


\bibitem{XuFact}
F.~Xu, Q.~Song and Y.~Q.~Jin,
``Polarimetric SAR image factorization,''
\emph{IEEE Trans. Geosci. Remote Sens.}, vol. 55, no. 9, pp. 5026-5041, Sep. 2017.


\bibitem{XuAnis}
F.~Xu, Y.~Li and Y.~Q.~Jin,
``Polarimetric-anisotropic decomposition and anisotropic entropies of high-resolution SAR images,''
\emph{IEEE Trans. Geosci. Remote Sens.}, vol. 54, no. 9, pp. 5467-5482, Sep. 2016.


\bibitem{IGARSS2019}
H.~Bi, F.~Xu, Z.~Wei, Y.~Han, Y.~Cui, Y.~Xue, and Z.~Xu,
``Unsupervised PolSAR Image Factorization with Deep Convolutional Networks,''
In \emph{Proc. IEEE IGARSS}, 2019, pp. 1061-1064.

\bibitem{Cao2020}
X.~Cao, J.~Yao, Z.~Xu, and D.~Meng,
``Hyperspectral Image Classification With Convolutional Neural Network and Active Learning,''
\emph{IEEE Trans. Geosci. Remote Sens.}, DOI:10.1109/TGRS.2020.2964627, 2020.


\bibitem{XuDeori2005}
F.~Xu, and Y.-Q. Jin,
``Deorientation theory of polarimetric scattering targets and application to terrain surface classification,''
\emph{IEEE Trans. Geosci. Remote Sens.}, vol. 43, no. 10, pp. 2351- 2364, Oct. 2005.


\bibitem{almm2019}
D.~Hong, N.~Yokoya, J.~Chanussot, and X.~X.~Zhu, ``An augmented linear mixing model to address spectral variability for hyperspectral unmixing,'' \emph{IEEE Trans. Image Process.}, vol. 28, no. 4, pp. 1923-1938, 2019.


\bibitem{DWT3D}
L.~C. Lin,
``A tutorial of the wavelet transform,''
NTUEE, Taiwan.


\bibitem{Mallat}
S.~G. Mallat,
``A theory for multiresolution signal decomposition: the wavelet representation,''
\emph{IEEE Trans. Pattern Anal. Mach. Intell.}, vol. 11, no. 7, pp. 674-693, 1989.

\bibitem{GraphBasedSemi}
M.~Culp, and G.~Michailidis,
``Graph based semi-supervised learning,''
\emph{IEEE Trans. Pattern Anal. Mach. Intell..}, vol. 30, no. 1, pp. 174-179, Jan. 2008.

\bibitem{JianSun}
J.~Sun and J. Ponce
``Learning Dictionary of Discriminative Part Detectors for Image Categorization and Cosegmentation,''
\emph{Int. J. Comput. Vis.}, vol. 120, no. 2, pp. 111-133, Nov. 2016.


\bibitem{Gibbs}
S.~Geman and D.~Geman,
``Stochastic relaxation, Gibbs distribution and
the Bayesian restoration of images,''
\emph{IEEE Trans. Pattern Anal. Mach. Intell.}, vol. 6, no. 6, pp. 721-741, Nov. 1984.


\bibitem{ICM}
R.~Szeliski, R.~Zabih, D.~Scharstein, O.~Veksler, V.~Kolmogorov,
A.~Agarwala, M.~Tappen, and C.~Rother,
``A Comparative Study of Energy Minimization Methods for Markov
Random Fields,''
In \emph{ECCV}, 2006.


\bibitem{BP2003}
M.~F. Tappen and W.~T. Freeman,
``Comparison of graph cuts with belief propagation for stereo, using identical MRF parameters,''
In \emph{ICCV}, 2003.


\bibitem{CXYCNN}
X.~Cao, F.~Zhou, L.~Xu, D.~Meng, Z.~Xu, and J.~Paisley,
``Hyperspectral Image Classification With Markov Random Fields and a Convolutional Neural Network,''
\emph{IEEE Trans. Image Process.}, vol. 27, no. 5, pp. 2354-2367, May 2018.


\bibitem{LIBLINEAR}
R.~E. Fan, K.~.W. Chang, C.~J. Hsieh, X.~R Wang and C.~J. Lin,
``LIBLINEAR: A library for large linear classification,''
\emph{J. Machine Learning Research}, vol.9, pp.1871-1874, Aug. 2008.


\bibitem{MLRsubMLL}
J.~Li, J.~M.~Bioucas-Dias, and A.~Plaza,
``Spectral-spatial hyperspectral image segmentation using subspace multinomial logistic regression and Markov random fields,''
\emph{IEEE Trans. Geosci. Remote Sens.}, vol. 50, no. 3, pp. 809-823, Mar. 2012.


\bibitem{CaoXiangyong}
X. Cao, L. Xu, D. Meng, Q. Zhao, and Z. Xu,
``Integration of 3-dimensional discrete wavelet transform and Markov random field for hyperspectral image classification,''
\emph{Neurocomputing}, vol. 226, pp. 90-100, Feb. 2017.



\end{thebibliography}
%

%
\vspace{-1cm}
\begin{IEEEbiography}[{\includegraphics[width=1in,height=1.25in,clip,keepaspectratio]{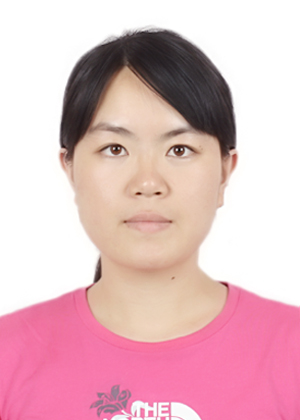}}]{Haixia Bi}
received the B.S. degree and M.S. degree in
computer science and technology from
Ocean University of China, Qingdao, China, in 2003 and 2006,
and the Ph.D. degree in computer science and technology
from Xi'an Jiaotong University, Xi'an, Jun. 2018.
She is currently a post-doctoral research fellow with
the University of Bristol, Bristol, United Kingdom.

Her current research interest lies in
machine learning, remote sensing image processing and big data.
\end{IEEEbiography}

\vspace{-1.2cm}
\begin{IEEEbiography}[{\includegraphics[width=1in,height=1.25in,clip,
keepaspectratio]{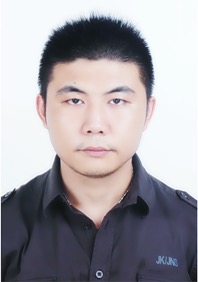}}]{Lin Xu}
	received his Ph.D. degree from Xi'an Jiaotong University. Upon  two years of academic experience as a postdoctoral researcher in Electrical and Computer Engineering at New York University Abu Dhabi and one-year industry experience as a senior researcher of the Institute of Advanced Artificial Intelligence in Horizon Robotics. He joined Shanghai Em-Data Technology Co., Ltd. as the director of the Artificial Intelligence Institute. He is currently working on machine learning models/algorithms, deep learning techniques, 2D/3D data analysis, and high-level computer vision recognition, e.g., classification, retrieval, segmentation, detection, and their applications in the security system, traffic control, and automatic drive.
His research interest lies in neural networks, learning algorithms, and applications in computer vision.
\end{IEEEbiography}

\vspace{-1.2cm}
\begin{IEEEbiography}[{\includegraphics[width=1in,height=1.25in,clip,
keepaspectratio]{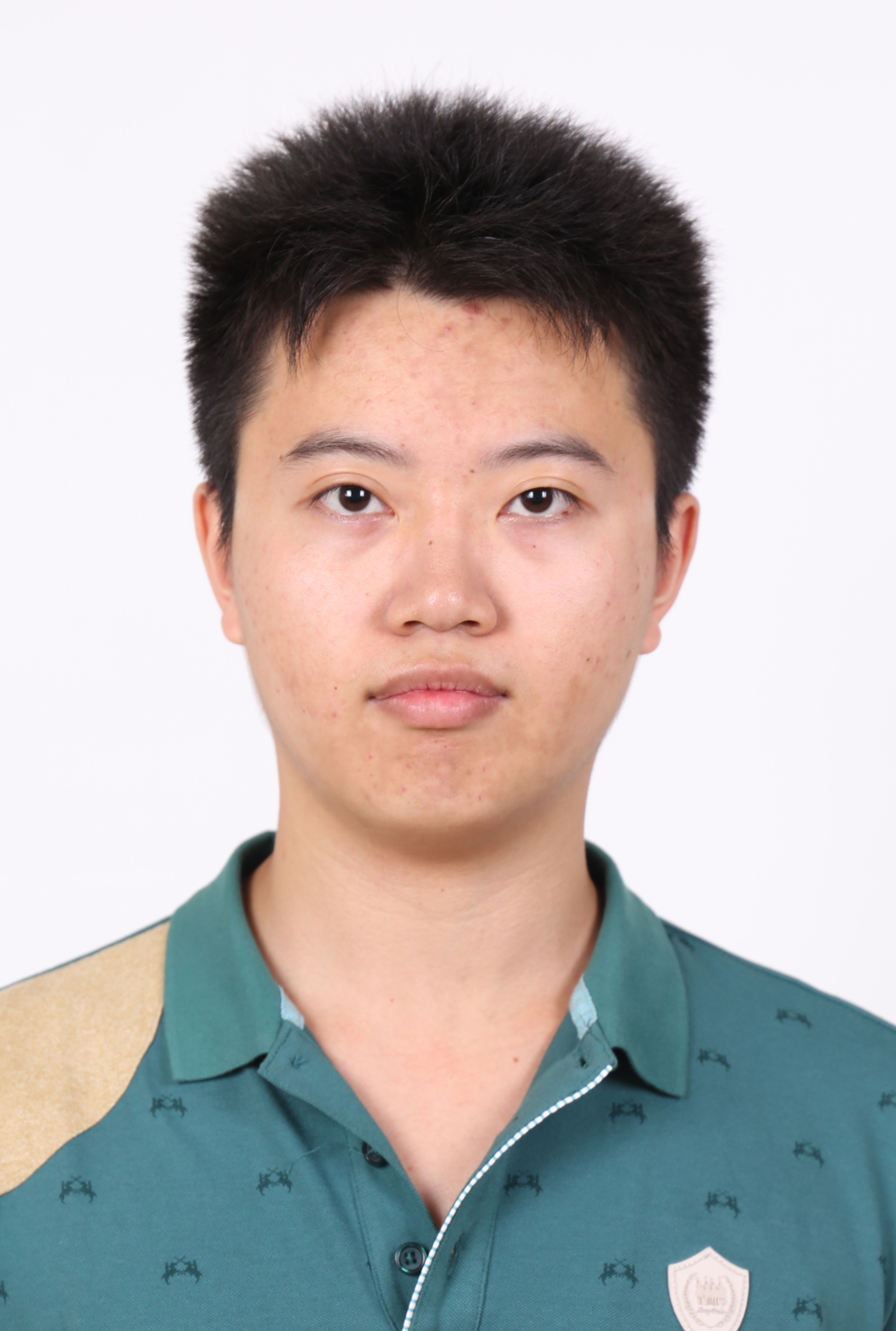}}]{Xiangyong Cao}
 received the B.Sc. and Ph.D degree from Xi'an Jiaotong University, Xi'an, China, in 2012 and 2018 respectively. He was a Visiting Scholar with Columbia University, New York, NY, USA, from 2016 to 2017.
 He is currently a post-doctoral research fellow with
Xi'an Jiaotong University, Xi'an, China.

 His current research interests include Bayesian method, deep learning, and hyperspectral image processing.
\end{IEEEbiography}

\vspace{-1.2cm}
\begin{IEEEbiography}[{\includegraphics[width=1in,height=1.25in,clip,keepaspectratio]{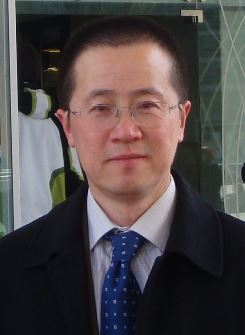}}]{Yong Xue}
(SM'04) received the B.Sc. degree in
physics and the M.Sc. degree in remote sensing
and GIS from Peking University, Beijing, China, in
1986 and 1989, respectively, and the Ph.D. degree
in remote sensing and GIS from the University of
Dundee, Dundee, U.K.
He is a Professor with the School of Electronics, Computing and Mathematics, University of Derby, Derby, United Kingdom.

His research
interests include middleware development for Geocomputation on spatial information grid, aerosol, digital earth, high throughput computation and telegeoprocessing.
Prof. Xue is editors of the \emph{International Journal of Remote Sensing},
\emph{International Journal of Digital Earth}, a chartered physicist, and
a member of the Institute of Physics, U.K.
\end{IEEEbiography}

\vspace{-1.2cm}
\begin{IEEEbiography}[{\includegraphics[width=1in,height=1.25in,clip,keepaspectratio]{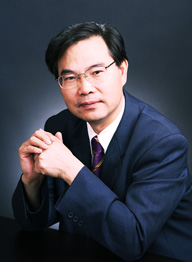}}]{Zongben Xu}
received the Ph.D. degree in mathematics
from Xi'an Jiaotong University, Xi'an, China,
in 1987.
He currently serves as the Chief Scientist of
the National Basic Research Program of China
(973 Project), and the Director of the Institute for
Information and System Sciences, Xi'an Jiaotong
University.
He is the Academician of the Chinese
Academy of Sciences. His current research interests
include intelligent information processing and
applied mathematics.

Prof. Xu was a recipient of the National Natural Science Award of China
in 2007, and the CSIAM Su Buchin Applied Mathematics Prize in 2008. He
delivered a 45 minute talk on the International Congress of Mathematicians
in 2010.
\end{IEEEbiography}






\end{document}